\documentclass[letterpaper, 10 pt, journal, twoside]{template/IEEEtran}
\IEEEoverridecommandlockouts                              % This command is only needed if 
\usepackage{graphicx}
\usepackage{caption, subcaption}
\usepackage{amsmath,amssymb,enumerate}

\usepackage{amsthm}
\usepackage{mathtools}
\usepackage{breqn}
\usepackage[usenames,dvipsnames,svgnames,table, xcdraw]{xcolor}
\usepackage[colorlinks=true,pdfpagemode=UseNone,citecolor=black,linkcolor=black,urlcolor=BrickRed]{hyperref}
\usepackage{algorithm, algpseudocode}

\usepackage{blindtext}
\usepackage{gensymb}
\usepackage{xparse}
\usepackage{lipsum}
\usepackage{mathrsfs}
\usepackage[mathscr]{euscript}
\usepackage{times}
\usepackage[noadjust]{cite} % to group references
\usepackage{multicol}
\usepackage{amsfonts}
\usepackage[utf8]{inputenc}
\usepackage[T1]{fontenc}
\usepackage{textcomp}
\usepackage{balance}
\usepackage{soul}
\usepackage{multirow}
\usepackage{booktabs}
\usepackage{sidecap}
\usepackage{makecell}
\usepackage{array,float}

\renewcommand{\thefigure}{\arabic{figure}}

\definecolor{free}{rgb}{0.9608, 0.5882, 0.3922}
\definecolor{building}{rgb}{0.0, 0.784, 1.0}
\definecolor{barrier}{rgb}{0.39, 0.16, 0.16}
\definecolor{other}{rgb}{0.22, 0.35, 0.31}
\definecolor{pedestrian}{rgb}{0.35, 0.12, 0.59}
\definecolor{pole}{rgb}{0, 0, 1}
\definecolor{road}{rgb}{1, 0, 1}
\definecolor{ground}{rgb}{0.31, 0.94, 0.59}
\definecolor{sidewalk}{rgb}{0.29, 0, 0.29}
\definecolor{vegetation}{rgb}{0, 0.69, 0}
\definecolor{vehicle}{rgb}{0.96, 0.59, 0.39}
\definecolor{fence}{rgb}{0.39, 0.16, 0.16}
\definecolor{sign}{rgb}{0, 0, 1}

\definecolor{scarColor}{rgb}{0.9608, 0.5882, 0.3922}
\definecolor{sbicycleColor}{rgb}{0.2608, 0.9020, 0.3922}
\definecolor{smotorcycleColor}{rgb}{0.5882, 0.2353, 0.1176}
\definecolor{struckColor}{rgb}{0.7059, 0.1176, 0.3137}
\definecolor{sothervehicleColor}{rgb}{1, 0.3137, 0.3922}
\definecolor{spersonColor}{rgb}{0.1176, 0.1176, 1}
\definecolor{sbicyclistColor}{rgb}{0.7843, 0.1569, 1}
\definecolor{smotorcyclistColor}{rgb}{0.3529, 0.1176, 0.5882}
\definecolor{sroadColor}{rgb}{0.5, 0, 1}
\definecolor{sparkingColor}{rgb}{1, 0.5882, 1}
\definecolor{ssidewalkColor}{rgb}{0.2941, 0, 0.2941}
\definecolor{sothergroundColor}{rgb}{0.2941, 0, 0.6863}
\definecolor{sbuildingColor}{rgb}{0, 0.7843, 1}
\definecolor{sfenceColor}{rgb}{0.1961, 0.4706, 1}
\definecolor{svegetationColor}{rgb}{0, 0.6863 ,0}
\definecolor{strunkColor}{rgb}{0, 0.2353, 0.5294}
\definecolor{sterrainColor}{rgb}{0.3137, 0.9412, 0.5882}
\definecolor{spoleColor}{rgb}{0.5882, 0.7412, 1}
\definecolor{strafficsignColor}{rgb}{0, 0, 1}

\newcommand{\m}{\mathop{\mathrm{m}}}

\title{\LARGE \bf ConvBKI: Real-Time Probabilistic Semantic Mapping Network \\ with Quantifiable Uncertainty}

\author{Joey Wilson, Yuewei Fu, Joshua Friesen, Parker Ewen\\
Andrew Capodieci, Paramsothy Jayakumar, Kira Barton, and Maani Ghaffari%
\thanks{DISTRIBUTION STATEMENT A. Approved for public release; distribution is unlimited. OPSEC\# 7836}
\thanks{J. Wilson, Y. Fu, J. Friesen, P. Ewen, K. Barton, and M. Ghaffari are with the University of Michigan, Ann Arbor, MI 48109, USA. {\tt\small{\{wilsoniv,ywfu,friesej\}@umich.edu}},
{\tt\small{\{pewen,bartonkl,maanigj\}@umich.edu}}}% <-this % stops a space
\thanks{A. Capodieci is with Neya Systems Division, Applied Research Associates, Warrendale, PA 15086, USA. {\tt\small{acapodieci@neyarobotics.com}}}% <-this % stops a space
\thanks{P. Jayakumar is with the US Army DEVCOM Ground Vehicle Systems Center, Warren, MI 48397, USA. {\tt\small{paramsothy.jayakumar.civ@army.mil}}}%
}

\begin{document}

\maketitle
\thispagestyle{empty}
\pagestyle{empty}

\begin{abstract}
In this paper, we develop a modular neural network for real-time {\color{black}(> 10 Hz)} semantic mapping in uncertain environments, which explicitly updates per-voxel probabilistic distributions within a neural network layer. Our approach combines the reliability of classical probabilistic algorithms with the performance and efficiency of modern neural networks. Although robotic perception is often divided between modern differentiable methods and classical explicit methods, a union of both is necessary for real-time and trustworthy performance. We introduce a novel Convolutional Bayesian Kernel Inference (ConvBKI) layer which incorporates semantic segmentation predictions online into a 3D map through a depthwise convolution layer by leveraging conjugate priors.
We compare ConvBKI against state-of-the-art deep learning approaches and probabilistic algorithms for mapping to evaluate reliability and performance.
We also create a Robot Operating System (ROS) package of ConvBKI and test it on real-world perceptually challenging off-road driving data.
\end{abstract}

% \begin{IEEEkeywords}
% Deep Learning for Visual Perception, Mapping, Semantic Scene Understanding. 
% \end{IEEEkeywords}

% \IEEEpeerreviewmaketitle

\section{Introduction}

Robotic perception is at a crossroads between classical, probabilistic methods and modern, implicit deep neural networks. Classical probabilistic solutions offer reliable and trustworthy performance at the expense of longer run-times and un-optimized, hand-tuned parameters. In contrast, modern deep learning methods achieve higher performance and improved latency due to optimization in an implicit space but lose the ability to generalize to new data. However, probabilistic and deep learning methods are not intrinsically opposed and may be combined through methodical design. 

Perception is the general study of understanding sensory information, and mapping is an important subset which seeks to consolidate sensory input into an informative world model. Maps are capable of representing high levels of scene understanding by storing multiple modalities of information, including semantic labels, dynamic motion, and topological relations. Although some works propose to discard a world model in favor of mapless end-to-end control \cite{MaplessMP3}, maps are still widely used due to their reliability and interpretable nature with shared human-robot scene understanding. Additionally, mapping within a robotic framework enables a modular schema for perception, planning, and control which avoids formulating overly complex and ill-posed problems. 

\begin{figure}[t]
    \centering
    \begin{subfigure}{\linewidth}
        \centering
        \includegraphics[width=\textwidth]{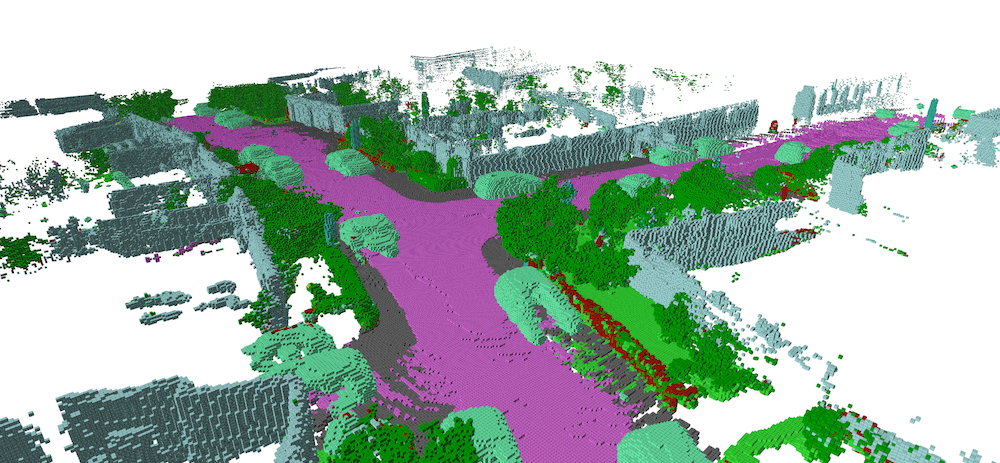}
        \caption{Semantic Categories}
        \label{fig:KITTI_Semantics}
    \end{subfigure}
    \begin{subfigure}{\linewidth}
        \centering
        \includegraphics[width=\textwidth]{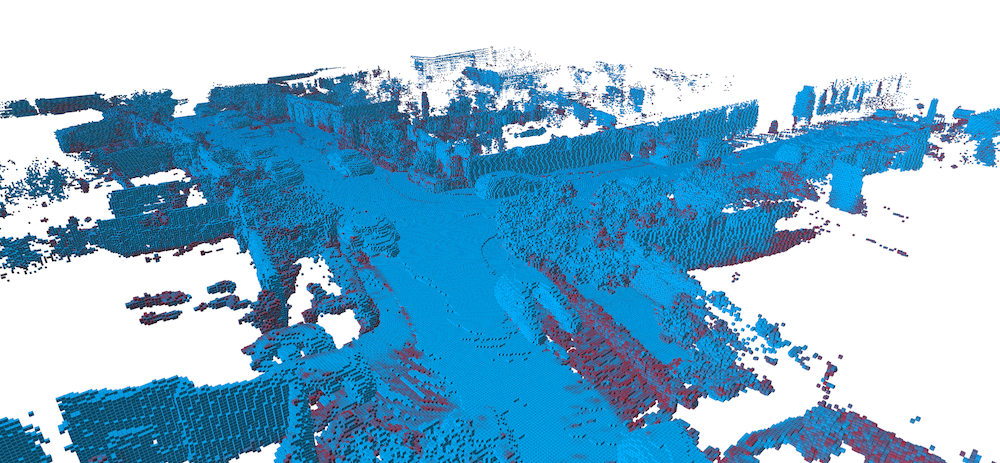}
        \caption{Variance}
        \label{fig:KITTI_Variance}
    \end{subfigure}
    \caption{Example output of our method (ConvBKI) on Semantic KITTI sequence 0 \cite{SemanticKITTI}. ConvBKI recurrently updates and maintains a semantic map with uncertainty in real-time by reframing the Bayesian update step as a depthwise separable convolution operation. Voxels encode the expected semantic category and variance as concentration parameters of the Dirichlet distribution, which is the conjugate prior of the categorical distribution and multinomial distribution. The input to the network is a 3D point cloud from a stereo-camera or LiDAR sensor. The points are labeled with semantic classification probabilities by a semantic segmentation network which provides per-point predictions. ConvBKI updates the semantic belief of each voxel by spatially weighting the input semantic point clouds through a unique geometric kernel learned for each semantic category. {\color{black}Fig. \ref{fig:KITTI_Semantics} shows the maximum likelihood semantic prediction per voxel, while Fig. \ref{fig:KITTI_Variance} shows the variance per voxel calculated with Eq. \eqref{eq:Variance}. Variance is linearly translated to RGB colors, where blue indicates voxels with a low variance and red indicates a high variance.}
    }
    \label{fig:Intro_Fig}
    \vspace{-4mm}
\end{figure}

One common mapping strategy is occupancy mapping which uses 3-dimensional cubic voxels in a grid-like structure to describe the environment with binary labels. While initial maps only contained binary free or occupied labels \cite{OccupancyGrid}, {\color{black}later works attempted to expand upon the scene understanding of mapping algorithms by incorporating semantic labels obtained by state-of-the-art semantic segmentation neural networks \cite{SemanticFusion, MappingSBKI}}. Semantic labels within the map can be used by downstream planners to complete tasks such as staying on a road or lifting a cup, instead of merely avoiding occupied space. Other work has also sought to improve the rigid, discrete structure of voxels with continuous geometry by surfels \cite{SumaPP, SemanticFusion}, signed distance functions \cite{PanopticTSDF}, meshes \cite{SemanticMesh, BKIMesh} and more, although these approaches can often be represented using occupancy maps \cite{Kimera}.

% These approaches often fail in the presence of dynamic objects which leave artifacts misleading to downstream planning tasks which rely upon accurate occupancy grids \cite{DynamicBKI, MotionSC}. 

In real-world operating conditions, robots must contend with uncertain inputs corrupted by noisy sensors and limited observations. In applications where safety is paramount, such as autonomous driving, it is vital to have a measure of certainty as well as algorithms robust to new, unseen data. For example, when driving off-road, mobile robots must contend with negative obstacles in addition to positive obstacles. Whereas positive obstacles are the presence of occupied space, a negative obstacle presents a unique challenge due to the lack of a driveable surface, such as a hole or cliff, which the mobile robot cannot traverse. Without uncertainty quantification, a mobile robot may incorrectly assert the absence of an obstacle or the presence of a road instead of slowing down and investigating before proceeding. 

Early approaches to probabilistic robotic mapping were based on the incorporation of input through Bayesian inference methods such as Gaussian Process Occupancy Maps \cite{GPOM}. However, Gaussian Processes have a cubic time complexity with respect to training samples due to expensive matrix inversion. Later Bayesian methods sought to improve the inference speed through methods such as Bayesian Generalized Kernel Inference \cite{BKIOccupancy}, which approximate distributions at model selection \cite{BKIProof} {\color{black}through kernel functions}. However, real-time operation is still a challenge for many hand-crafted probabilistic mapping approaches. 

In contrast, recent mapping approaches have established end-to-end world modeling architectures where high levels of scene understanding are stored in a hidden state and updated recurrently across time \cite{MotionNet,  ImageToMap, SemanticMapNet}. While implicit maps are able to achieve high performance with limited latency due to efficient and parallel modeling, they still encounter training difficulties and challenges when exposed to new data outside of their training set. Additionally, supervised approaches face the challenge of obtaining data sets which can be particularly challenging for dynamic mapping \cite{MotionSC, DynamicBKI}. 

In this paper, we combine probabilistic and differentiable mapping approaches to obtain a real-time {\color{black}($> 10$ Hz)} 3D semantic mapping network which balances the efficiency of deep learning approaches with the reliability of classical methods by operating on explicit probability distributions in a modular, parallelized architecture. We demonstrate that ConvBKI is able to perform comparably with deep learning approaches on data sets it has been trained on and transfer more readily to unseen data sets, including a new, challenging off-road driving data set.

\subsection{Contributions} 
We pose the Bayesian semantic mapping problem as a differentiable neural network layer (ConvBKI) in order to achieve the accelerated inference rates and optimized performance of deep learning while maintaining the ability to quantify uncertainty in closed form. {\color{black}Compared to previous semantic mapping works such as Semantic Fusion \cite{SemanticFusion} and Semantic BKI \cite{MappingSBKI}, ConvBKI also performs semantic mapping in a dense voxel space with Bayesian methods but reformulates the map inference as an end-to-end neural network with differentiable parameters.} We demonstrate the improved performance and inference rates compared to probabilistic mapping algorithms and demonstrate reliability compared to the most similar learning-based comparison, Semantic MapNet \cite{SemanticMapNet}. We also present real-world results in an open-source ROS node on challenging off-road data. 

This paper is built upon our previous work \cite{MotionSC}, where we established a new data set and neural network for semantic mapping in dynamic scenes {\color{black}called MotionSC}. However, MotionSC was unable to bridge the sim-to-real gap from the CarlaSC simulated data set to real-world Semantic KITTI \cite{SemanticKITTI} and required accurate 3D ground truth maps, which are often unavailable in the real world. 

{\color{black}Compared to our previous conference paper on Convolutional Bayesian Kernel Inference \cite{ConvBKI}, we accelerate the mapping algorithm by modifying the voxel storage to a sliding local window which discards voxels outside of the local boundaries. Additionally, we add studies
on the reliability of ConvBKI which compare ConvBKI with end-to-end mapping network Semantic MapNet on the sim-to-real transfer between Carla SC and Semantic KITTI. We also add more in-depth ablation studies on the effect of dynamic objects, noisy input, and the run-time of local mapping compared to global map storage. Lastly, we gather a real-world perceptually challenging off-road dataset and study the ability of ConvBKI to transfer from off-road driving dataset RELLIS-3D \cite{Rellis3D} to our new dataset, which is {\color{black}unrepresented by} the training set.} 

In summary, our contributions are:

\begin{enumerate}[i.]
    \item Novel neural network layer for closed-form Bayesian inference, which combines the best of probabilistic and differentiable programming.
    \item Real-time explicit probabilistic incorporation of semantic segmentation predictions into a map with quantifiable uncertainty. 
    \item Comparison of the proposed ConvBKI approach against the state-of-the-art learning-based and probabilistic semantic mapping methods.
    % \item Open-source software as a ROS package for ease of use with testing on a new off-road dataset, available at \href{https://github.com/UMich-CURLY/BKI_ROS}{https://github.com/UMich-CURLY/BKI_ROS}. 
\end{enumerate}

\subsection{Outline} The remaining content of this article is organized as follows. Section \ref{sec:Background} provides a literature review on the evolution of mapping algorithms and off-road perception. Section \ref{sec:Preliminaries} provides an overview of Bayesian Kernel Inference necessary to understand ConvBKI. Section \ref{sec:ConvBKI} proposes a new deep learning layer for real-time incorporation of semantic segmentation predictions within a map, trained only on semantic segmentation data. Section \ref{sec:Local Mapping} accelerates the framework with local mapping. Section \ref{sec:Results} compares ConvBKI against learning and probabilistic approaches in addition to ablation studies for the network design. Section \ref{sec:offroad} concludes with a ROS package for open-source use and evaluates ConvBKI on challenging real-world off-road driving data. {\color{black}The ROS package can be found at \href{https://github.com/UMich-CURLY/BKI_ROS}{https://github.com/UMich-CURLY/BKI_ROS} with examples on sequences of data from Semantic KITTI \cite{SemanticKITTI} and RELLIS-3D \cite{Rellis3D}. The new off-road data is unable to be released publicly due to proprietary reasons.} Finally, limitations and possibilities for future work are discussed in Section \ref{sec:Conclusion}.

% Whereas traditional robotics algorithms sought mathematically founded approaches to perception and controls, many modern approaches disregard domain expertise in favor of large black-box approaches with quick run-times and high accuracy. However, by operating in a lower dimensional latent space without concrete mathematical foundations, such black-box algorithms are vulnerable to failure when exposed to data outside of the training set. Therefore, a middle-ground for robotic perception is necessary which balances the robust mathematical foundations of traditional methods with the efficiency and precision of modern algorithms. 

% How we propose to do it

% In this paper, we introduce a high definition semantic mapping framework for dynamic environments which runs as an end-to-end neural network while still performing explicit physical operations at each step. 
% % Mention each section

% Contributions
% ConvBKI layer
% Attention layer
% Results
% ROS package

% \jw{First page, top right: example output of our map on Neya data. \\ Top of second page will be an entire row showing our work transferring vs. learned approaches transferring to new data.}
\section{Literature Review}\label{sec:Background}
{\color{black}In this section, we review 3D semantic mapping and the trade-offs between learned and hand-crafted approaches. We also review approaches for off-road perception, and the importance of uncertainty in perceptually challenging environments.}

\subsection{Semantic Mapping}
% Kimera
Semantic mapping is the task of incorporating semantic labels such as car or road into a geometric world model. Historically, most mapping methods were hand-crafted and mathematically derived. Early semantic mapping algorithms semantically labeled images, then projected to 3D and directly updated voxels through a voting scheme or Bayesian update \cite{SemanticBayes3D, SemanticFrequent3D, SemanticFusion, SemanticVoting3D}. Later semantic mapping algorithms applied further optimization through Conditional Random Fields (CRF), which encourages consistency between adjacent voxels \cite{SemanticCRF, SemanticCRF2, SemanticCRF3}. Separately, continuous mapping algorithms estimate occupancy through a continuous non-parametric function such as Gaussian processes (GPs) \cite{OccupancyGP, GPOM}. However, these methods suffer from a high computation load, rendering them impractical for onboard robotics with large amounts of data. For example, GPs have a cubic computational cost with respect to the number of data points and semantic classes \cite{SemanticGP}. {\color{black}Bayesian Kernel Inference is a method proposed to accelerate continuous mapping by approximating GPs at model selection, which will be discussed in more detail in Section \ref{sec:Preliminaries}.}

Other works have also explored alternative data-efficient representations such as surfels \cite{SumaPP}, truncated signed distance functions \cite{PanopticTSDF}, and meshes \cite{SemanticMesh, Kimera}, although these methods are not exclusive of semantic mapping and can be used as an extension to create high-fidelity semantic maps \cite{Hydra, Kimera}. While in this work we focus on real-time semantic mapping, our algorithm can be extended to create real-time semantic meshes through methods such as VoxBlox \cite{VoxBlox}. 

Many modern approaches to mapping leverage inventions in the deep learning community to learn an efficient, implicit approximation of the world in a lower dimensional latent space. Transformers are large arrays of multi-head attention layers \cite{Attention} which can be combined to form large language models \cite{BERT}, vision transformers \cite{VisionTransformer}, {\color{black}and networks for 3D object detection \cite{DeformableTransformer} and mapping \cite{ImageToMap} among other applications}. However, transformers are notoriously difficult to train and suffer un-diagnosable failures when exposed to new data. Other modern approaches to semantic mapping apply recurrent neural networks \cite{RecurrentOctoMap, DA-RNN, SemanticMapNet} or spatiotemporal convolution networks \cite{MotionNet, MotionSC} to model spatiotemporal dynamics. Spatiotemporal approaches remove the recurrency of other mapping algorithms and instead aggregate past information to operate directly on a finite length temporal dimension of information. While operating on a finite length time dimension is efficient for convolutional networks, the assumption discards valuable memory. Additionally, learning approaches for mapping require large amounts of supervised data which can be challenging to obtain in real-world dynamic driving environments \cite{SemanticKITTI}.  

% Other recent works have explored approximating continuous geometry implicitly with Neural Radiance Fields (NeRF) \cite{BlockNerf, NeRF} or occupancy networks \cite{OccupancyNetworks, NeuralBlox, ConvOccupancy}, in order to negate the expensive memory of voxels.

Although learning-based approaches have achieved success in minimizing memory and accelerating inference, they still encounter significant challenges. By approximating functions implicitly, there is no notion of when a network will fail, as provided by variance, or the ability to diagnose an error. Additionally, latent space operations are vulnerable to errors when exposed to new data, motivating research into problems such as the sim-to-real gap. On the other side of the spectrum, mathematical hand-crafted approaches provide reliability and trustworthiness at the cost of efficiency. Our approach seeks to find a middle ground by operating on explicit probability distributions in an end-to-end differentiable neural network with modern deep learning layers. {\color{black}Next, we discuss literature in off-road perception which necessitates a balance between efficiency and reliability.}

\subsection{Off-Road Perception}
Perception in off-road driving is challenging due to less structured environments with more ambiguity and unbalanced data than on-road driving. Roads generally contain a finite set of easily distinguishable classes with a regular geometric pattern, such as the location and shape of roads, vehicles, traffic lights, etc. However, off-road presents more irregular geometric shapes which are not easily segmented and depend on context \cite{Rellis3D}. Although on-road perception has advanced tremendously due to a plethora of large-scale datasets \cite{SemanticKITTI, Waymo_Cite, NuScenes}, fewer off-road semantic segmentation datasets are available \cite{Rellis3D, RUGD, Yamaha}. RELLIS-3D is the largest off-road LiDAR semantic segmentation data set at the time of writing, containing 13,556 labeled point clouds. The labels are unequally distributed, dominated by grass, tree, and bush semantic categories. Uncertainty-aware semantic segmentation \cite{KPConv} significantly outperforms baselines \cite{SalsaNext} on RELLIS-3D \cite{Rellis3D}, emphasizing the importance of uncertainty quantification on challenging unstructured outdoor perception. 

Other works have found that geometric information alone is insufficient for off-road autonomy due to ambiguous structure \cite{DARPA, DARPA_CMU, Off-Road-Obstacle}, necessitating more information on traversability through semantic mapping \cite{Yamaha}. For example, without semantic context, a bush or tall blade of grass may be confused as impenetrable obstacles. Due to these challenges of off-road navigation, we test the semantic mapping and uncertainty quantification of ConvBKI on a new off-road data set {\color{black}in Section \ref{sec:Results}. In the next section we introduce background on efficient probabilistic mapping necessary to understand our proposed method.} 
{\color{black}\section{Preliminaries}\label{sec:Preliminaries}
% In this section we introduce preliminaries on the Bayesian Kernel Inference (BKI) mapping method which our work extends.
Semantic Bayesian Kernel Inference (S-BKI) \cite{MappingSBKI} is a 3D continuous semantic mapping framework which builds on the work of Vega-Brown et al. \cite{BKIProof} and Doherty et al. \cite{BKIOccupancy}. Semantic BKI recursively incorporates semantically labeled points into a voxelized map by considering the relative position between each input point and voxel through a kernel function. Bayesian Kernel Inference (BKI) is an efficient approximation of GPs which leverages local kernel estimation to reduce computation complexity from $\mathcal{O}(N^3)$ operations to $\mathcal{O}(\text{log} N)$, where $N$ is the number points. In contexts such as mapping, there may be hundreds of thousands of points, rendering GPs impractical.

\begin{figure}[t]
    \includegraphics[width=0.7\linewidth]{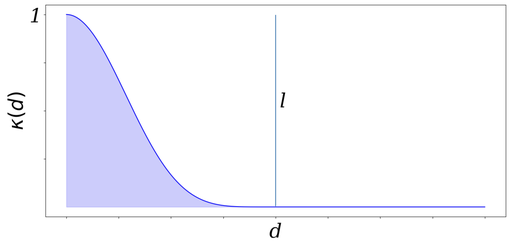}
    \centering
    \caption{Sparse Kernel Function. $\kappa(d)$ has a maximum value of 1 at $d=0$, and decays to 0 by $d=l$. When applied to semantic mapping, points proximal to the voxel centroid have more influence over the semantic label of the voxel than points further from the centroid.
    }
    \label{fig:SparseKernel}
    \vspace{-4mm}
\end{figure}

For supervised learning problems, our goal is to identify the relationship $p(y_*|x_*,\mathcal{D})$ from a sequence of $N$ independent observations $\mathcal{D} = \{(x_1, y_1), ..., (x_N, y_N)\}$ for query point $x_*$ with label $y_*$. In semantic mapping, the input data is a series of semantically labeled points with positions $x_i$ and semantic observations $y_i,$ and $*$ represents the query voxel. Observations are associated with latent model parameters $\theta_i$ which define the likelihood $p(y_i | \theta_i)$. In the case of semantic mapping, we wish to infer the categorical parameters $\theta_i$, such that the probability of observing a one-hot encoded semantic category $y_i$ may be written as: 

\begin{equation}\label{eq:likelihood_categorical}
    p(y_i | \theta_i) = \prod_{c = 1}^C (\theta_i^c)^{y_i^c} .
\end{equation}

\noindent As we observe 3D semantic point-pairs $(x_i, y_i)$, our goal is to learn the latent target parameters $\theta_*$ which define the Categorical posterior of each voxel $*$ with centroid $x_*$. Note that the voxel posterior $p(\theta_* | x_*, \mathcal{D})$ is dependent on observations continuously scattered around the voxel, requiring an update which considers the relative position of input data.

% \citeauthor{BKIProof}
Vega-Brown et al.~\cite{BKIProof} propose a solution to relate the extended likelihood  $g(y_i) = p(y_i | \theta_*, x_i, x_*)$ to the likelihood distribution $f(y_i) = p(y_i | \theta_*)$ through a positional kernel $k(x_i, x_*)$. This relationship generalizes local kernel estimation to Bayesian inference for supervised learning. In particular, they relate the extended likelihood and likelihood by showing that the maximum entropy distribution $g$ satisfying $D_{KL}(g||f) \geq \rho(x_*, x_i)$ has the form $g(y_i) \propto f(y_i) ^ {k(x_*, x_i)}$, where $k$ is a kernel function and $D_{KL}$ is the Kullback-Leibler Divergence. In this case, {\color{black}$\rho: X \times X \to \mathbb{R}^+$} is some function which bounds information divergence between the likelihood distribution $f(y_i) = p(y_i | \theta_*)$ and the extended likelihood distribution $g(y_i) = p(y_i | \theta_*, x_i, x_*)$. Concretely,

\begin{equation}\label{eq:bki}
    p(y_i | x_i, \theta_*, x_*) \propto p(y_i | \theta_*)^{k(x_i, x_*)} .
\end{equation}

\noindent Functions $k$ and $\rho$ have an equivalence relationship, where each is uniquely determined by the other. The only requirements are that 
{\color{black}
\begin{equation}\label{eq:req1}
    k(x, x) = 1 \forall x \quad \text{and} \quad k(x, x_*) \in [0, 1] \forall x, x_*.
\end{equation}
}

As described above, in semantic mapping $\theta_*$ describes a Categorical distribution over the semantic categories $C$. Therefore, the extended likelihood spatially relates semantic observations $(x_i, y_i)$ to voxel latent parameters $\theta_*$ using a kernel function $k$. This formulation is especially useful for likelihoods $p(y_i|\theta_*)$ chosen from the exponential family, as the likelihood raised to the power of $k(x_*, x)$ is still within the exponential family. 

% \citeauthor{BKIOccupancy} 
 Doherty et al.~\cite{BKIOccupancy} then apply the BKI kernel model to the task of occupancy mapping. In occupancy mapping, occupied points are measured by a 3D sensor such as LiDAR, and free space samples can be approximated through ray tracing. Measurement $x_i \in \mathbb{R}^3$ represents a 3D position with corresponding observation $y_i^c \in \{0, 1\}$ {\color{black}where ($y_i^1 = 1$) for free space measurements and ($y_i^2 = 1$) for occupied measurements at the end of a ray.} {\color{black}The voxel occupancy can be modeled by a Bernoulli distribution over $\theta_*$.} Adopting the conjugate prior distribution Beta($\alpha_*^{1}, \alpha_*^{2}$) over voxel parameters $\theta_*$ yields a closed-form update equation by applying Bayes' rule: 
 
\begin{equation}\label{eq:Bayes Update}
    p(\theta_* | x_*, \mathcal{D}) \propto p(\mathcal{D} | \theta_*, x_*) p(\theta_* | x_*),
\end{equation}
which can be written in terms of the extended likelihood as

\begin{equation}\label{eq:Posterior Extended Likelihood}
    p(\theta_* | x_*, \mathcal{D}) \propto \left[ \prod_{i=1}^N p(y_i | x_*, x_i, \theta_*) \right] p(\theta_* | x_*).
\end{equation}

Applying the relationship between extended likelihood and likelihood defined in Eq. \eqref{eq:bki}, the posterior can be simplified to

\begin{equation}\label{eq:Posterior Continuous}
    p(\theta_* | x_*, \mathcal{D}) \propto \left[ \prod_{i=1}^N p(y_i | \theta_*) ^{k(x_i, x_*)} \right] p(\theta_* | x_*).
\end{equation}

Due to the conjugate relationship between the prior $p(\theta_*|x_*)$ and likelihood $p(y_i | \theta_*)$ defined in Eq. \eqref{eq:likelihood_categorical}, the update yields a closed-form solution where:

\begin{equation}\label{eq:Posterior Likelihood}
    p(\theta_* | x_*, \mathcal{D}) \propto \left[ \prod_{i=1}^N \left[ \prod_{c = 1}^C (\theta_*^c)^{y_i^c} \right]^{k(x_i, x_*)} \right] \prod_{c = 1}^C (\theta_*^c)^{\alpha_*^c - 1
    },
\end{equation}
which simplifies to 

\begin{equation}\label{eq:Posterior Simple}
    p(\theta_* | x_*, \mathcal{D}) \propto \prod_{c = 1}^C (\theta_*^c)^{\alpha_*^c + \sum_{i=1}^N y_i^c k(x_i, x_*) -1}.
\end{equation}

Since the posterior is proportional to the conjugate prior distribution Beta($\alpha_*$), the update at time-step $t$ for voxel $*$ can be written in closed-form as
\begin{equation}\label{eq:Occupancy Update}
    \alpha^c_{*,t} = \alpha^c_{*, t-1} + \sum_{i=1}^{N_t}k(x_*, x_i) y_i^c.
\end{equation}

The equation provides a closed-form method for updating the belief that voxel * is occupied or free given observed measurements and samples of free space. {\color{black}The un-normalized posterior Beta($\alpha_{*,t}^{1}, \alpha_{*, t}^{2}$) is a weighted summation of nearby inputs points, where influence of nearby points is calculated through the kernel function. From the un-normalized voxel concentration parameters $\alpha_*$, the normalized expectation and variance of voxel parameters $\theta_*$ are defined as

\begin{equation}\label{eq:Variance}
    \eta_* = \sum_{j=1}^C \alpha^j_*, \quad \mathbb{E}[\theta^c_*] = \frac{\alpha^c_*}{\eta_*}, \quad \mathbb{V}[\theta^c_*] = \frac{
    \frac{\alpha^c_*}{\eta_*} (1 - \frac{\alpha^c_*}{\eta_*}) 
    }{ 1 + \eta_*
    }.
\end{equation}}

{\color{black}
Doherty et al.~\cite{BKIOccupancy} propose to use a sparse kernel function \cite{SparseKernel} which simplifies kernel calculation to a function of the Euclidean distance between two points. For two points $x_i$ and $x_*$, the sparse kernel is calculated as
}

{\color{black}\begin{align}
\label{eq:SparseKernel}
        \nonumber &\kappa(d) = \\ 
        &\begin{cases}
          \sigma_0 [\frac{1}{3} (2 + \cos (2 \pi \frac{d}{l})(1 - \frac{d}{l}) +
          \frac{1}{2 \pi} \sin (2 \pi \frac{d}{l})) ], 
          & \text{if } d < l \\
          0, & \text{else}
        \end{cases}
\end{align}
where $d := \lVert x_i - x_* \rVert_2$ and $l$ is the kernel length. The variables $\sigma_0$ is used to scale the magnitude of the kernel, however $\sigma_0 = 1$ to satisfy Eq. \ref{eq:req1}.} {\color{black} Note that whereas the kernel function $k: \mathbb{R}^3 \times \mathbb{R}^3 \to [0, 1]$ is defined between two points, the sparse kernel function $\kappa: \mathbb{R}^3 \to [0, 1]$ is a function of the Euclidean distance between the two points. The sparse kernel is shown in {\color{black}Figure} \ref{fig:SparseKernel}, and effectively provides more weight to close points.}

%\citeauthor{MappingSBKI} 
Gan et al.~\cite{MappingSBKI} show that the same approach can be applied to semantic segmentation predictions from neural networks by adopting a Categorical likelihood over $\theta_*$ with conjugate prior Dir($C, \alpha_0$). {This model is also calculated using Eq. \eqref{eq:Occupancy Update}, however the number of classes has increased from two to any positive integer $C$.}

Although Semantic BKI has achieved accurate and efficient performance in 3D mapping compared to previous methods, it is still limited in key ways. Firstly, the kernel is hand-crafted, and kernel parameters must be manually tuned. As a result, a single spherical kernel is shared between all semantic classes. Second, the update operation is computationally expensive, as the kernel inference must locate all nearby voxels and does not leverage the parallelization of modern GPU architectures. In Section \ref{sec:ConvBKI} we show how ConvBKI improves upon these short-comings.}

% Introduce the layer
% Demonstrate results
% What is the issue with the layer?
\section{Convolutional Bayesian Kernel Inference}\label{sec:ConvBKI}
\begin{figure}[t]
    \includegraphics[width=0.95\linewidth]{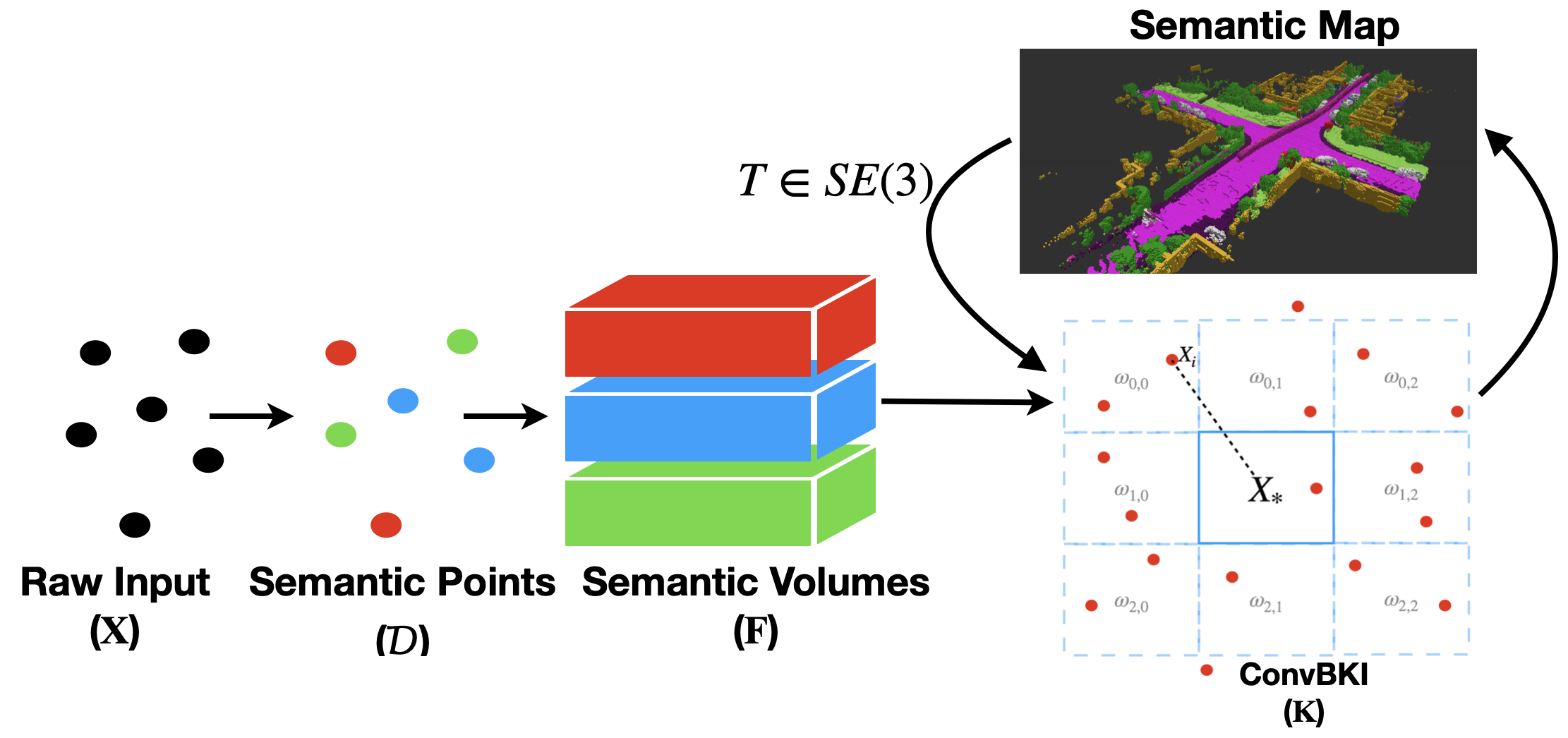}
    \centering
    \caption{Structural overview of ConvBKI. Input 3D points are labeled with semantic segmentation predictions by a semantic segmentation neural network, and are voxelized into a dense voxel grid ($\textbf{F}$) where each voxel contains the summation over the semantic segmentation predictions of all points within the voxel. A 3D depthwise separable convolution performs the Bayesian update in real-time, where the posterior semantic voxel grid is equal to the convolution of $\textbf{F}$ plus the prior voxels from the last time-step. 
    }
    \label{fig:Diagram}
    \vspace{-4mm}
\end{figure}

In this section, we propose a novel neural network layer, Convolutional Bayesian Kernel Inference (ConvBKI), intended to accelerate and optimize S-BKI. Compared to S-BKI, ConvBKI \emph{learns} a unique kernel for each semantic class and generalizes to 3D ellipsoids instead of restricting distributions to spheres. We introduce the layer and incorporate it into an end-to-end deep neural network for updating semantic maps. A figure summarizing our approach is shown in Fig. \ref{fig:Diagram}.

\subsection{Layer Definition} 
We build a faster, trainable version of Semantic BKI based on the key observation that Eq. \eqref{eq:Occupancy Update} may be rewritten as a depthwise convolution \cite{MobileNet}. We find that the kernel parameters are differentiable with respect to a loss function and, therefore, may be optimized with gradient descent. Learning the kernel parameters enables more expressive geometric-semantic distributions and improved semantic mapping performance. 

{\color{black}First, we note that semantic segmentation neural networks predict soft-max encoded predictions, and one-hot encoding the measurement $y$ loses information. Intuitively, the closed-form update is a weighted semantic counting model which should be able to consider soft-max observations from segmentation neural networks. Therefore, we propose to consider the likelihood of soft-max prediction $y_i$ as

\begin{equation}\label{eq:likelihood_continuous}
    p(y_i | \theta_i) \propto \prod_{c = 1}^C (\theta_i^c)^{y_i^c}, 
\end{equation}
where $\sum_{c=1}^C y_i^c = 1$ and $y_i^c \geq 0$. Substitution into Eq. \eqref{eq:Posterior Likelihood} yields the same closed-form update as Eq. \eqref{eq:Occupancy Update}, and enables consideration of one-hot encoded or soft-max encoded neural network observations. The distribution defined in Eq. \eqref{eq:likelihood_continuous} is a specific case of the Continuous Categorical distribution proposed in \cite{ContinuousCategorical}, and has also been noted as useful for the BKI framework in \cite{EvidentialBKI}.}

% Whereas Semantic BKI shares a single kernel between all semantic categories, we propose a unique kernel for each semantic class and a compound kernel capable of learning ellipsoid distributions instead of only spherical distributions. 

% Add references 
% VoxBlox, Chimera

The update operation in Eq. \eqref{eq:Occupancy Update} performs a weighted sum of semantic probabilities over the local neighborhood of voxel centroid $x_*$. This operation can be directly interpreted in continuous space with radius neighborhood operations such as in PointNet++ \cite{PointNet++}, DGCNN \cite{EdgeConv}, or KPConv \cite{KPConv}. However, in practice these operations are much too slow to compute for hundreds of thousands of camera or LiDAR points due to an expensive k-Nearest Neighbor operation. Instead, we perform a discretized update, where the geometric position of each local point is approximated by voxelized coordinates. Approximation through downsampling is already performed in Semantic BKI \cite{MappingSBKI}, and is a common step in real-time mapping literature \cite{VoxBlox, Kimera}. 

% \Parker{I dont know what this means}

{\color{black}Our method updates a dense voxel grid of a fixed size at each time step, with voxel concentration parameters initialized to prior at the first time step. For the prior, we use a small positive value such $\alpha_*^c = \epsilon$ for all voxels and semantic categories. We apply a discretized update to the dense voxel grid of dimension $\mathbb{R}^{D_C \times D_X \times D_Y \times D_Z}$, where $D$ represents the dimension of the semantic channel ($C$) and Euclidean ($X, Y, Z$) axes. 

\begin{figure}[t]
    \centering
    
    \begin{subfigure}[t]{0.2\textwidth}
        \centering
        \includegraphics[width=\textwidth]{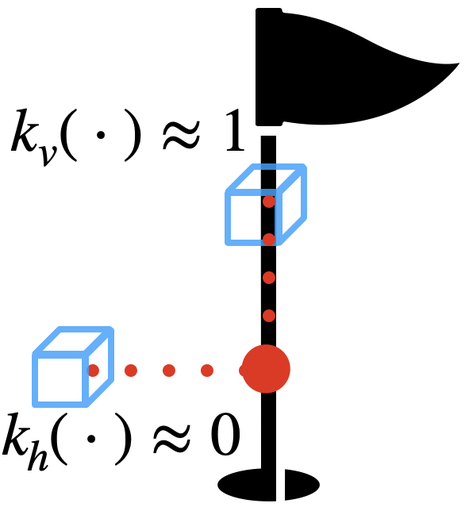}
        \caption{Pole Intuition}
        \label{fig:pole_voxels}
    \end{subfigure}
    \hfill
    \begin{subfigure}[t]{0.23\textwidth}
        \centering
        \includegraphics[width=\textwidth]{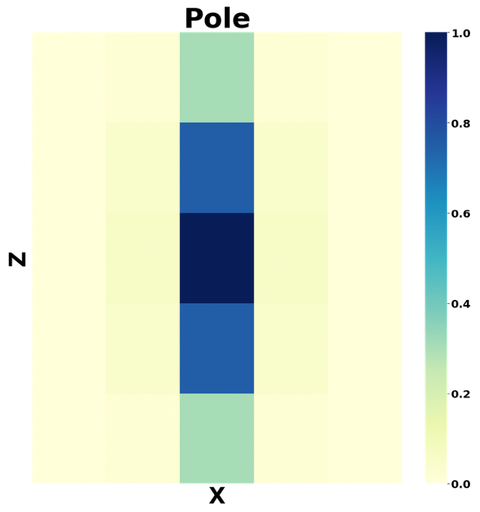}
        \caption{Learned Weights}
        \label{fig:pc_road1}
    \end{subfigure}
    \caption{Illustration of compound kernel motivation. ConvBKI learns a distribution to geometrically associate points with voxels. Whereas a point (red) labeled pole suggests a vertically adjacent voxel (blue) may also be a pole, it does not imply the same for a horizontally adjacent voxel. The image on the right demonstrates a slice of the learned weights for the class pole, which matches the expectations shown on the left. 
    }
    \label{fig:PoleExample}
    \vspace{-4mm}
\end{figure}

For each input point cloud, we discretize input points to voxelized coordinates and create a dense semantic volume $\textbf{F},$ where input $\textbf{F}_*^c$ is the sum of the probability mass for semantic category $c$ of all predictions $y_i$ contained in voxel $*$. Let $I(*, i)$ be an indicator function representing whether point $x_i$ lies within voxel $*$. We compute input semantic volume $\textbf{F} \in \mathbb{R}^{D_C \times D_X \times D_Y \times D_Z}$ as follows.}

\begin{equation}\label{eq:Input Definition}
    \textbf{F}^c_{*} =  \sum_{i=1}^{N} I(*, i) y_i^c 
\end{equation}
%\Parker{Is the kernel centered at *?}
For each voxel, the Bayesian update can be calculated as the sum of the prior and a depthwise convolution over input $\textbf{F}${\color{black}, equivalent to a discretized representation of Eq. \eqref{eq:Occupancy Update}}. Let $h, i, j$ be the discretized coordinates of voxel $*$ within $\textbf{F}$, and {\color{black} $n, o, p$ be indices within convolution filter $\mathbf{K} \in \mathbb{R}^{D_C \times f \times f \times f}$ where $f$ is the filter size. Then, we can write the update for a single semantic channel of voxel $*$ as} 
{\color{black}
\begin{equation}\label{eq:Depthwise Update}
    \alpha^c_{*, t} = \alpha^c_{*, t-1} + \sum_{n, o, p}  \textbf{K}^c_{n, o, p} \textbf{F}^c_{h+n, i+o, j+p},
\end{equation}
where integer indices $n, o, p \in [-\frac{f-1}{2}, \frac{f-1}{2}]$.} Note that this is the equation for a zero-padded depthwise convolution, where dense 3D convolution is performed at each voxel in the feature map, with a unique convolution filter $\textbf{K}^c$ per semantic category $c$. As a result, this operation may be accelerated by GPUs and optimized through gradient descent since the convolution operation is differentiable. 

{\color{black} While it is possible to learn an individual weight for each position and semantic category in filter $\textbf{K}$, we found that restricting the number of parameters through a kernel function increases the ability of the network to quickly learn generalizable semantic-geometric distributions. Following \cite{MappingSBKI} and \cite{BKIOccupancy}, we use a sparse kernel \cite{SparseKernel} as our kernel function since the sparse kernel fulfills the requirements listed in {\color{black}Eq. \eqref{eq:req1}}. Additionally, the sparse kernel is differentiable so that a partial derivative of the loss function with respect to the kernel parameters can be calculated. A plot of the sparse kernel function is included in Fig. \ref{fig:SparseKernel} for reference and is described mathematically in Eq. \eqref{eq:SparseKernel}, where the parameters are kernel length $l$, and signal variance $\sigma_0$. Note that for Eq. \eqref{eq:req1} to remain valid, $\sigma_0$ must be 1, leaving only one tune-able parameter for the kernel function. }

%\Parker{You dont explain what $K^c$ represents, and from how you define it I think it should either be $K_{c,k,l,m,}$ or define $K^c$ rather than $K$}

Effectively, kernel $\mathbf{K}$ is a weight matrix where each weight represents a semantic and spatial likelihood of points being correlated around the query voxel. {\color{black}Therefore, we propose to learn a unique kernel $k^c$ for each semantic category and assign convolutional filter weights to $\mathbf{K}$ by discretizing the kernel function. For a filter of dimension $f$ and resolution $\Delta r$, the convolution parameters $\textbf{K}^c_{n,o,p}$ at filter indices $n, o, p$ can be calculated by evaluating kernel function $k^c$ at the offset of position $n, o, p$ relative to the filter center as follows.}

% For example, if an input point is labeled road, then voxels nearby along the $X$ or $Y$ axes are also likely to be road and would have a large weight close to 1. In contrast, a point labeled as pole would have more influence over voxels nearby vertically rather than horizontally. An example slice of the kernel is illustrated in Fig. \ref{fig:KernelDemo}, which shows that the optimized kernels match our intuition. After training on Semantic KITTI \cite{SemanticKITTI}, kernels converge such that the kernel for {\color{black} the pole category has a large weight across the vertical ($Z$) axis, while the kernel for the road category has a large weight along the horizontal axes.} 

% \Parker{Maybe define this as $k^c$ to be consistent with later notation?}

{\color{black}
\begin{equation}\label{eq:KernelWeight}
    \textbf{K}^c_{n, o, p} = k^c \left(\textbf{0}, 
    \Delta r \left( \frac{f-1}{2} - 
    \begin{bmatrix}
        n \\
        o \\
        p
    \end{bmatrix} \right) \right)
\end{equation}
}

% \Parker{You can try using the "\\left" and "\\right" bracket operators in the above equations, dont know if itll look good though.}
To accommodate complex geometric structures of real objects, we also propose a compound kernel~\cite[Ch. 4]{rasmussen2006gaussian} computed as the product of a {\color{black}sparse kernel over the horizontal plane ($\kappa_h$) and vertical axis ($\kappa_v$).} \color{black}{The compound kernel is defined as
\begin{equation}\label{eq:CompoundKernel}
\small 
    k^c\left(\begin{bmatrix}
        x_i^1\\
        x_i^2\\
        x_i^3\\
    \end{bmatrix} ,
    \begin{bmatrix}
        x_*^1\\
        x_*^2\\
        x_*^3\\
    \end{bmatrix}\right)
    = \kappa_h^c\left(\left\|
    \begin{bmatrix}
        x_i^1 - x_*^1 \\
        x_i^2 - x_*^2\\
    \end{bmatrix}\right\|_2\right) 
    \kappa_v^c\left(\left\|
        x_i^3 - x_*^3
    \right\|_2\right), 
\end{equation}
}
\noindent{\color{black}where the integer superscript represents the Euclidean axes.} 

Intuitively, ConvBKI treats the output of a semantic segmentation neural network as sensor input and learns a geometric probability distribution over each semantic class. Semantic classes have different shapes, where classes such as poles are more vertical, and classes such as road have influence horizontally. This idea is visualized in Fig. \ref{fig:PoleExample}, where points vertically adjacent to a pole have a higher weight than points horizontally adjacent to a pole. 

{\color{black}Beyond semantic labels, some important considerations for applying ConvBKI include free space measurements and dynamic environments. In our experiments, we focus on semantic quality and do not incorporate free space observations. Instead, we filter free space voxels through sampling by only visualizing voxels with $\eta_* > 0.1$, where a voxel with a measurement would have $\eta_* \geq 1$. In applications where explicit free space is desired, explicit free space measurements can be incorporated into the semantic map by sampling along each ray at fixed distance intervals \cite{BKIOccupancy}, and treating the sampled free space points as another semantic category when performing the update \cite{MappingSBKI}. Free space sampling is incorporated into our software, and inference results of free-space sampling are included in Section \ref{Sec:Ablation}.

Another important detail to note is that ConvBKI does not draw distinction between static and dynamic objects, leaving artifacts in the map as objects move as traces regardless of how free space is calculated \cite{DynamicBKI}. This concept is demonstrated in Section \ref{Sec: Neural Comparison}, where we purposefully evaluate ConvBKI in highly dynamic scenes without any special consideration paid to dynamic objects. Methods exist which propose to either separately track dynamic objects from the static world, or use velocity information to decay the occupied belief \cite{DynamicBKI}. }

\subsection{Training}
One key advantage of ConvBKI over alternative learning-based mapping approaches is the cost of data. ConvBKI seeks to find the maximum entropy distribution $g$, which spatially relates query point $i$ with voxel $*$ as $g(y_i) = p(y_i | \theta_*, x_i, x_*$). Therefore, the network is learning a function to improve semantic segmentation through the distribution $g$ recurrently.
As a result, ConvBKI only requires semantic segmentation ground truth as opposed to a 3D map since the learning problem is framed as spatially and temporally integrating semantic predictions to best perform semantic segmentation of point $x_i$.
In contrast, other mapping approaches rely upon a ground truth semantic map, which is difficult to obtain in the real world \cite{SemanticMapNet, MotionSC}. 

% \Parker{As an alternative to the previous paragraph?}
% Key advantages of ConvBKI over alternative learning-based mapping approaches is the reduced cost of data and the flexibility of ConvBKI to be integrated as a post-processing layer to pre-trained networks.
% ConvBKI learns the spatial distribution of semantic classes and allows it to act as a Bayes filter to semantic segmentation network outputs.

To train ConvBKI, we formulate the learning problem as a semantic segmentation task given the past $\mathcal{T}$ frames of semantic point clouds labeled using an off-the-shelf semantic segmentation network.
3D points are labeled by an off-the-shelf semantic segmentation network to form $\mathcal{D}_{t-\mathcal{T}:t}$, then transformed to the current frame $T_t$.

{\color{black}Next, a map of size $\mathbb{R}^{D_C \times D_X \times D_Y \times D_Z}$ is initialized to the prior distribution. The prior distribution assigns a small positive value $\epsilon$ to each Dirichlet parameter $\alpha$ so that each voxel is initialized with a high variance and uniform expected value following Eq. \eqref{eq:Variance}. Experimentally, we set $\epsilon = 1e^{-6}$ to enforce a weak prior.
Points with semantic labels are then grouped into voxels to form discretized measurement $\textbf{F}_t$ through Eq. \eqref{eq:Input Definition}.
Finally, the ConvBKI filter is applied to calculate the posterior distribution per-voxel. The filter is trained as a maximum-likelihood estimator by maximizing the likelihood of each point in the final input point cloud. Note that the likelihood of each semantic category $c$ per voxel $*$ is calculated using Eq. \eqref{eq:Variance} which normalizes the probability of each semantic category, $\mathbb{E}[\theta^c_*]$, by the sum of the un-normalized concentration parameters, $\eta^c_*$. Since ConvBKI directly outputs semantic likelihoods, it is trained using the negative log-likelihood loss instead of cross-entropy loss \cite{PyTorch}.}
\section{Local Mapping}\label{sec:Local Mapping}
\begin{figure}[t]
    \centering
    \includegraphics[width=0.6\linewidth]{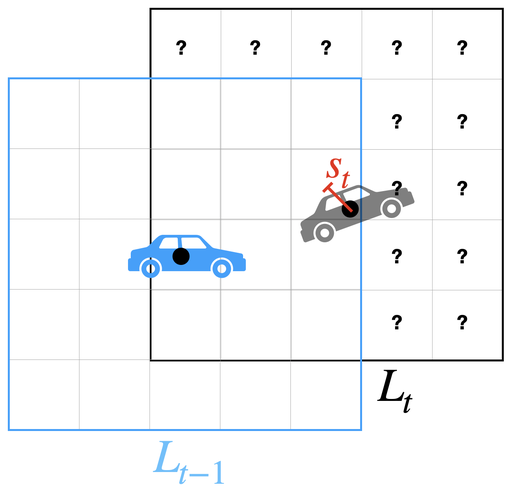}
    \caption{{\color{black}Toy 2D illustration of local mapping method. When the ego vehicle moves from time $t-1$ to a new location at time $t$, the ego-vehicle position is approximated to the nearest voxel centroid. The local map $L_{t-1}$ shifts to the new map center and discards voxels outside of $L_t$. New voxels are initialized to prior, represented by the question mark. Since the center of map is discretized, an offset represented by $s_t$ in red is maintained to transform input point clouds to the map frame. This method prioritizes local semantic scene understanding over loop closure, as voxels will be initialized to prior when re-visited.}} 
\label{fig:LocalMap}
\vspace{-4mm}
\end{figure}

The input to ConvBKI at each time step is a point cloud voxelized into a dense local map $L_t$ with  semantic predictions $\textbf{F}_t$.  As the number of voxels in the semantic map increases with time, the cost of constructing $L_t$ grows exponentially. While ConvBKI may run at speeds of over 100 Hz depending on the size of the map, the bottleneck quickly becomes the query operation from the global map. Therefore, we propose to maintain a dense local map in memory and discard voxels outside of the local region similar to \cite{Hydra, KinectFusion}. In practice, a dense world model may be maintained in memory and integrated with a lower definition global model \cite{MapLite}, similar to planning through a local and global planner. {\color{black}Additionally, simultaneous localization and mapping (SLAM) algorithms have already achieved high inference rates and accuracy \cite{LIOSAM}, so we prioritize local semantic scene understanding over loop closure which may result from global maps.} An illustration of our local mapping approach is shown in Fig. \ref{fig:LocalMap}. 

{\color{black}While transforming the local map to match the continuous rotation and translation of the ego-robot is an intuitive approach, it requires approximations of the concentration parameters. We propose to exactly shift the local map by multiples of the voxel resolution, while rotating and translating the input point cloud data to map the local map frame.} 
3D input may be easily rotated and translated to match the local map coordinate frame without approximations, leading to lossless propagation.
Lossless propagation is especially important for explicit probability distributions, as compounding approximations lead to blurring of the map. 
% The disadvantage, however, is extra overhead as query points must be transformed from ego-centric coordinates to map-centric points when making predictions.
This approach is applied in real-world robots such as quadrupeds \cite{AnyMALGridMap} and is shown in Fig. \ref{fig:LocalMap}.

{\color{black}
Our approach consists of several simple steps, optimized for minimal overhead.
At each time-step a new pose $T_t \in \text{SE(3)}$ is provided by a localization algorithm.
From this pose, we calculate the transformation to the starting pose $T_0$:
\begin{equation}
    T_{t, 0} = T_0^{-1} T_t.
\end{equation}
% as well as from the previous pose to the starting pose, $T_{t-1, 0}$.
\noindent Each relative transformation $T_{t, 0}$ is composed of a rotation $R_{t} \in \text{SO(3)}$ and a translation $\psi_t \in \mathbb{R}^3$.
The translation between time $t-1$ and $t$ is then calculated in voxel coordinates, dependent on voxel resolution $\Delta r$:
\begin{equation}
    \nu_{t, t-1} = \left \lfloor \frac{\psi_{t}}{\Delta r} \right \rceil - \left \lfloor \frac{\psi_{t-1}}{\Delta r} \right \rceil ,
\end{equation}
where $\left \lfloor \cdot \right \rceil$ indicates the floating point value is rounded to the nearest integer.
Last, we allocate a new local map $\hat{L}_{t}$ of the same shape as $L_{t-1}$ and copy voxel data within the boundaries of $\hat{L}_{t}$ from $L_{t-1}$ with an index offset of $\nu_{t,t-1}$.

Since the local map was translated by an integer multiple of the voxel resolution, there may be an offset and rotation from ego coordinates to map coordinates. Therefore, the transformation from sensor coordinates to map coordinates is composed of rotation $R_t$ and an offset 

\begin{equation}
    s_t = \psi_t - \Delta r \left \lfloor \frac{\psi_{t}}{\Delta r} \right \rceil .
\end{equation}
} 

Intuitively, the translation in voxels of the local map is identified by rounding the translation from the initial coordinate frame to the current ego-centric coordinate frame.
Since the local map translation is rounded to an integer, an offset $s_t$ must be maintained to transform input point clouds from the ego-centric frame to the local map frame. Points must also be rotated by $R_t$ to match the orientation of the initial position. This approach requires minimal computational overhead and removes any approximation to ensure reliability.

\section{Results}\label{sec:Results}
In this section, we evaluate ConvBKI against probabilistic and learning-based approaches on several data sets {\color{black}with both camera and LiDAR sensors}. We also present ablation studies demonstrating the effects of design choices and visualizing the intuitive kernels learned by ConvBKI.

For all results, we train ConvBKI with a learning rate of 0.007 using the Adam optimizer \cite{ADAM} for one epoch with the weighted negative log-likelihood loss. We initialize the kernel length parameter for each kernel to $l = 0.5 \m$ and train with the last $\tau = 10$ frames as input. The kernel size is set to $f = 5$ with a resolution of $0.2 \m$. 

Inference rates are reported for the dense voxel update step of ConvBKI on an NVIDIA RTX 3090 GPU. Latencies are measured for a batch size of 1 and a single frame, averaged over $1000$ repetitions. Note that the update inference rate is directly correlated with the size of the voxel grid, and the map inference rate also depends on free space sampling and voxel grid construction which are studied in Sec. \ref{Sec:Ablation}. However, even including the latency of the semantic segmentation network and a voxel grid with resolution $0.1 \m$, ConvBKI maintains real-time inference rates with frequencies greater than $10~\mathrm{Hz}$. 

\subsection{Probabilistic Comparison}
\begin{table}[b]
\vspace{-4mm}
% \vspace{2mm}
% \footnotesize
\centering
\caption{Results on KITTI Odometry sequence 15 \cite{KITTI_Seq_15}. {\color{black}As the degree of kernel expressivity increases, mIoU of ConvBKI also improves from 76.5\% without optimization to 77.7\% with fully optimized per-class compound kernels.}}
\resizebox{0.48\textwidth}{!}{
\begin{tabular}{l|cccccccc|c|c}
% \toprule
% \multicolumn{3}{c|}{Method}
\multicolumn{1}{l}{\bf Method}&

\cellcolor{building}\rotatebox{90}{\color{white}Building} &
\cellcolor{road}\rotatebox{90}{\color{white}Road} &
\cellcolor{vegetation}\rotatebox{90}{\color{white}Vege.} &
\cellcolor{sidewalk}\rotatebox{90}{\color{white}Sidewalk} & 
\cellcolor{vehicle}\rotatebox{90}{\color{white}Car} & 
\cellcolor{pole}\rotatebox{90}{\color{white}Sign} &
\cellcolor{barrier}\rotatebox{90}{\color{white}Fence} &
\cellcolor{pole}\rotatebox{90}{\color{white}Pole} &
\rotatebox{90}{\bf mIoU (\%)} &
\rotatebox{90}{\bf Freq. (Hz)}\\ 

\hline 

\vspace{-2mm} \\
Segmentation \cite{Deep_Dilated_CNN} & 92.1 & 93.9 & 90.7 & 81.9 & 94.6 & 19.8 & 78.9 & 49.3 & 75.1\\
\bottomrule
\vspace{-2mm} \\
Yang et al. \cite{YangMethod9} & \textbf{95.6} & 90.4 & \textbf{92.8} & 70.0 & 94.4 & 0.1 & \textbf{84.5} & 49.5 & 72.2\\
BGKOctoMap-CRF \cite{BKIOccupancy} & 94.7 & 93.8 & 90.2 & 81.1 & 92.9 & 0.0 & 78.0 & 49.7 & 72.5\\
S-CSM \cite{MappingSBKI} & 94.4 & 95.4 & 90.7 & 84.5 & 95.0 & 22.2 & 79.3 & 51.6 & 76.6\\
S-BKI \textcolor{black}{(fine)} & 94.6 & 95.4 & 90.4 & 84.2 & 95.1 & \textbf{27.1} & 79.3 & 51.3 & 77.2 & 0.6\\

\bottomrule
\vspace{-2mm} \\
\textcolor{black}{S-BKI (0.2m)} & 92.6 & 94.7 & 90.9 & 84.5 & 95.1 & 21.9 & 80.0 & 52.0 & 76.5\\
ConvBKI Single & 92.7 & 94.8 & 90.9 & 84.7 & 95.1 & 22.1 & 80.2 & 52.1 & 76.6\\
ConvBKI Per Class & 94.0 & 95.5 & 91.0 & 87.0 & 95.1 & 22.8 & 81.8 & 52.9 & 77.5\\
ConvBKI Compound & 94.0 & \textbf{95.6} & 91.0 & \textbf{87.2} & \textbf{95.1} & 22.8 & 81.9 & \textbf{54.3} & \textbf{77.7} & \textbf{44.3}\\
\end{tabular}
}

\label{tab:kitti_iou}
% \vspace{-4mm}
\end{table}
First, we compare our network to its most direct probabilistic comparisons on the KITTI Odometry dataset \cite{KITTI_Odometry} {\color{black}with semantic labels obtained from \cite{KITTI_Seq_15}} following the approach of Semantic BKI \cite{MappingSBKI}. The purpose of comparing ConvBKI against probabilistic baselines is to highlight the quantitative advantages of kernel optimization, as well as the accelerated inference rate. 

The test set contains a sequence of {\color{black}one hundred consecutive} raw RGB images with depth estimated by ELAS \cite{ELAS_Depth} and pose estimated by ORB-SLAM \cite{ORBSLAM}. Semantic segmentation predictions for the image were generated by the deep network dilated CNN \cite{Deep_Dilated_CNN}. {\color{black}Ground truth semantic segmentation labels are available for twenty-five of the one hundred input images test set images from \cite{KITTI_Seq_15}.} Each baseline recurrently generates a global map from the stream of time-synchronized pose, depth, and semantic segmentation predictions. The goal of each algorithm is to estimate semantic segmentation labels at the current time step, given all past information. Therefore, baselines are compared by projecting pixels with depth to 3D world coordinates and assigning semantic segmentation to each pixel corresponding to the voxel the 3D projection resides in. Semantic predictions are aggregated over all time-steps in a global semantic map, and mapping accuracy is evaluated by the Intersection over Union (IoU) metric. 

Baselines include a CRF-based semantic mapping system \cite{YangMethod9}, BGKOctoMap-CRF \cite{BKIOccupancy, MappingSBKI}, S-BKI \cite{MappingSBKI} and Semantic-Counting Sensor Model (S-CSM) \cite{MappingSBKI}. Our most direct comparison is S-BKI, as the goal of our network is to accelerate and optimize the algorithm in a deep learning framework. However, {\color{black}S-BKI uses a resolution of 0.1 meters with point downsampling instead of discretization. In contrast, our method discretizes input point clouds into voxels to simplify the kernel calculation as a convolution.} Therefore, to directly evaluate the benefit of optimization, we add a version of S-BKI with a resolution of 0.2 meters and discretization, which we label S-BKI ($0.2 \m$). Grid bounds for ConvBKI are set to a range of [-40, -40, -5.0] to [40, 40, 5.0] $\m$. Results are summarized in Table \ref{tab:kitti_iou}.

\textbf{Kernel Optimization:} The results of Table \ref{tab:kitti_iou} demonstrate the utility of optimizing kernels, as subsequently more expressive kernels achieve higher mIoU. Starting from S-BKI (0.2m) which represents an unoptimized version of ConvBKI with a single filter shared between all classes, performance is slightly better than the segmentation input. Even without optimization, the BKI framework is capable of recurrently integrating information to improve semantic segmentation. With optimization over a single kernel shared by all classes, denoted ConvBKI Single, performance slightly increases from $76.5$ to $76.6 \%$ mIoU. The small increase is due to only a single parameter being optimized, yielding an un-expressive kernel. When a kernel is optimized for each semantic category (ConvBKI Per Class), mIoU increases substantially from $76.6$ to $77.5\%$. Finally, dividing the kernel for each semantic category into the product of a horizontal and vertical kernel (ConvBKI Compound) again increases performance, this time most notably for the pole class. Intuitively, points vertically adjacent to points on a pole are also likely to be a pole, whereas horizontally adjacent points are not. This idea is supported in Section \ref{Sec:Ablation}, which visualizes the kernels learned by ConvBKI Compound. 

\begin{figure*}[t]
    \centering
    
    \begin{subfigure}[t]{0.23\textwidth}
        \centering
        \includegraphics[width=\textwidth]{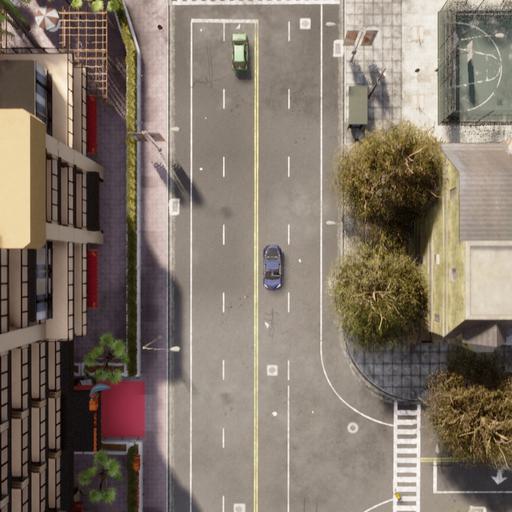}
        \caption{BEV Camera}
        \label{fig:bev_gt}
    \end{subfigure}
    \hfill
    \begin{subfigure}[t]{0.23\textwidth}
        \centering
        \includegraphics[width=\textwidth]{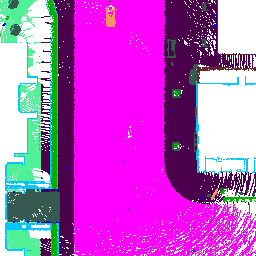}
        \caption{BEV Ground Truth}
        \label{fig:gt_label}
    \end{subfigure}
    \hfill
    \begin{subfigure}[t]{0.23\textwidth}
        \centering
        \includegraphics[width=\textwidth]{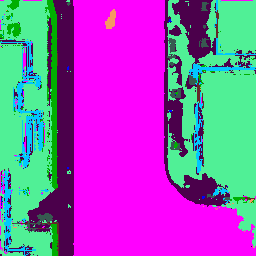}
        \caption{Semantic MapNet (GRU)}
        \label{fig:smnet_pred}
    \end{subfigure}
        \hfill
    \begin{subfigure}[t]{0.23\textwidth}
        \centering
        \includegraphics[width=\textwidth]{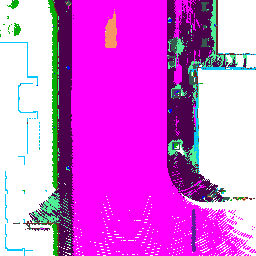}
        \caption{ConvBKI}
        \label{fig:convbki}
    \end{subfigure}
    \caption{Comparison of ConvBKI and Semantic MapNet on BEV mapping in CarlaSC. Both networks correctly segment the driveable surface and buildings. Compared with Semantic MapNet, ConvBKI does not make predictions on unseen regions, creating some sparsity in the map. ConvBKI correctly identifies the pedestrian in the bottom right, however leaves a trace as the pedestrian walks along the crosswalk. 
    In contrast, Semantic MapNet omits the pedestrian to achieve a higher quantitative performance.} 
    \label{fig:CarlaSC_qualitative}
\vspace{-4mm}
\end{figure*}

\textbf{Latency:} Another notable comparison in Table \ref{tab:kitti_iou} is between S-BKI (fine) and ConvBKI Compound. S-BKI (fine) uses a finer resolution of $0.1$ m with {\color{black}downsampling instead of discretization of the input point cloud. However, by leveraging differentiability, ConvBKI Compound is able to improve semantic mapping performance to surpass that of S-BKI. Additionally, ConvBKI achieves significant acceleration, with an inference frequency of 44.3 Hz compared to 2 Hz for S-BKI with downsampling and 0.6 Hz without. In summary, ConvBKI achieves an accelerated inference rate and optimized performance compared to probabilistic baselines.}

\subsection{Neural Network Comparison}\label{Sec: Neural Comparison}
Next, we compare ConvBKI Compound against several types of memory for recurrent neural networks on the task of Bird's Eye View (BEV) semantic mapping. The purpose of this evaluation is to isolate the trade-offs of ConvBKI opposed to more expressive algorithms with thousands of times more parameters. Specifically, we compare the ground-truth data used, memory, quantitative performance, inference rate, and reliability when transferring to a different dataset. 

In contrast to the global mapping comparison for probabilistic mapping, we evaluate on local BEV semantic mapping for a learning comparison. BEV mapping is more common for learning-based algorithms due to the memory cost of 3D operations, and provides a measure of scene understanding by evaluating the semantic map of ConvBKI from a different persepective.

\begin{table}[b]
\vspace{-4mm}
    \centering
    \caption{Comparison with learning-based memory on simulated data. {\color{black}Metrics are reported for the isolated recurrent networks, excluding the input segmentation network.}}
    \begin{tabular}{r||c|c|c|c|}
        {\it Method} & GRU & LSTM & Linear & ConvBKI  \\
        \bottomrule
        \bottomrule 
        \vspace{-2mm} \\
        {\it Map Dimension} & 2 & 2 & 2 & \textbf{3}  \\
        \bottomrule
        \vspace{-2mm} \\
        {\it Memory} (MB) & 2765 & 2917 & 2707 & \textbf{2497}  \\
        \bottomrule
        \vspace{-2mm} \\
        {\it Parameters} & 1.96 M & 2.04 M & 1.74 M & \textbf{22}  \\
        \bottomrule
        \vspace{-2mm} \\
        {\it mIoU} (1 epoch) & 24.0 & 25.3 & 24.7 & \textbf{26.7} \\
        \bottomrule
        \vspace{-2mm} \\
        {\it mIoU} (15 epoch) & \textbf{35.4} & 30.9 & 32.6 & N/A  \\
        \bottomrule
        \vspace{-2mm} \\
        {\it Freq.} (Hz) & 111.0 & 107.4 & 116.8 & \textbf{176.6}  \\
        \bottomrule
        \vspace{-2mm} \\
        {\it Data} & BEV & BEV & BEV & \textbf{Segmentation}  \\
        \bottomrule
    \end{tabular}
    \label{tab:CarlaResults}
\end{table}

We chose to compare with Semantic MapNet \cite{SemanticMapNet} as it was the most direct open-source learning-based comparison we were able to find. Semantic MapNet is a BEV semantic mapping algorithm which uses features from the last layer of a pre-trained semantic segmentation algorithm as input to per-cell recurrent units which perform a map update in feature space. Semantic MapNet groups points with features from the semantic segmentation network into square cells and applies a BEV filter which discards all but the highest point in each cell. Each cell with an input point is then updated through a learned memory model, including a Gated Recurrent Unit (GRU) \cite{GRU}, Long Short-Term Memory (LSTM) \cite{LSTM} or a linear layer. At inference time, a series of convolution layers are applied to the latent-space memory to produce a BEV semantic segmentation prediction for each 2-dimensional cell. 

{\color{black}Semantic MapNet functions similarly to ConvBKI except for several key differences. First, Semantic MapNet operates in two dimensions using only the highest point in each map cell as input, compared to the 3D representation of ConvBKI which incorporates every input point. Additionally, Semantic MapNet applies a recurrent network which updates each cell using latent features of the input points, whereas ConvBKI applies an explicit Bayesian update on the semantic predictions of the semantic segmentation network.}

The input to each ConvBKI and Semantic MapNet is from the same semantic segmentation network, Sparse Point-Voxel Convolutional Neural Network (SPVCNN) \cite{SPVNAS}. SPVCNN is an efficient semantic segmentation for LiDAR point clouds which leverages sparsity to accelerate inference. We follow the training procedure from Semantic MapNet with two modifications, including changing the number of input frames to $\tau = 10$ to match ConvBKI and the batch size to 2 in order to reduce training memory. Semantic MapNet was trained for fifteen epochs over the course of three days. For evaluation, each network is provided the past $\tau = 10$ frames as input and evaluated on the accuracy of the BEV local semantic map produced. Following the evaluation procedure of Semantic MapNet, cells with no input are discarded from evaluation. 

% We re-train SPVCNN on CarlaSC and use the publicly available weights for Semantic KITTI.  Weights for the SPVCNN are frozen before training the mapping networks. Whereas ConvBKI is trained to incorporate past information to best predict semantic segmentation labels, Semantic MapNet is trained directly on BEV mapping using the ground truth semantic labels obtained from CarlaSC. In the real world, ground-truth BEV maps are difficult to obtain. 

\subsubsection{Simulated Data}
First, we train ConvBKI and Semantic MapNet on the simulated driving dataset CarlaSC gathered from the CARLA simulator \cite{MotionSC}. CarlaSC contains complete 3D scenes with ground truth semantic labels for each voxel at a resolution of $0.2$ m and a sensor configuration designed to mimic the popular real-world Semantic KITTI \cite{SemanticKITTI} driving dataset. {\color{black}Additionally, CarlaSC is a highly dynamic dataset with moving actors which we use to highlight the artifacts left behind by ConvBKI as discussed earlier in Section \ref{sec:ConvBKI}.}

Ground truth BEV semantic maps are obtained by ray-tracing within each ground truth 3D map until an occupied voxel is found, and assigning the BEV semantic label to match the first occupied voxel. If no occupied voxel is encountered, the 2D cell is ignored during training and evaluation. We compare ConvBKI and Semantic MapNet on the test set of CarlaSC, which includes three scenes of 3 minutes in length sampled at 10 Hz. We use the remapped CarlaSC labels which contain 11 semantic categories: free space, building, barrier, other, pedestrian, pole, road, ground, sidewalk, vegetation and vehicles. Models are evaluated on the mIoU of all semantic predictions in the 5,400 frame test-set. Ground truth scenes are at a resolution of 0.2 meters with local boundaries with respect to the on-board LiDAR sensor of (-25.6, -25.6, -2.0) meters to (25.6, 25.6, 1.0) meters. 

\begin{figure}[b]
    \includegraphics[width=0.8\linewidth]{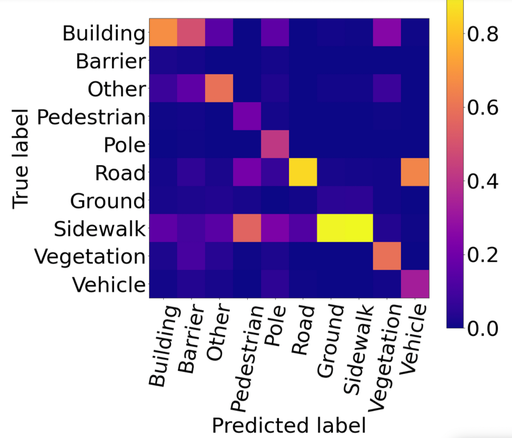}
    \centering
    \caption{Confusion matrix of ConvBKI compound on simulated BEV mapping, normalized over the predictions. ConvBKI leaves traces from dynamic objects, causing sidewalk to be incorrectly labeled {\color{black}{as}} pedestrian and road as vehicle.  
    }
    \label{fig:confusion_matrix}
\end{figure}

\begin{figure*}[t]
    \centering
    \begin{subfigure}[t]{0.3\textwidth}
        \centering
        \includegraphics[width=\textwidth]{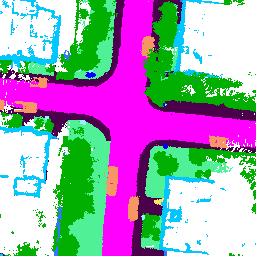}
        \caption{Ground Truth}
        \label{fig:gt_k}
    \end{subfigure}
    \hfill
    \begin{subfigure}[t]{0.3\textwidth}
        \centering
        \includegraphics[width=\textwidth]{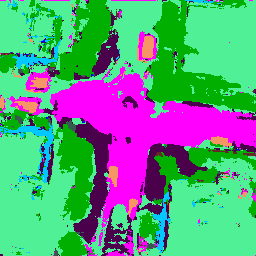}
        \caption{Semantic MapNet Zero-Shot}
        \label{fig:sm_k}
    \end{subfigure}
    \hfill
    \begin{subfigure}[t]{0.3\textwidth}
        \centering
        \includegraphics[width=\textwidth]{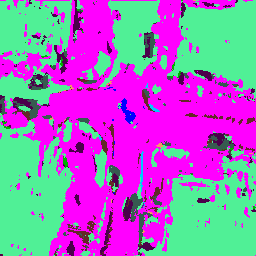}
        \caption{Semantic MapNet Re-trained Segmentation}
        \label{fig:sm_k_t}
    \end{subfigure}
    \hfill
    \begin{subfigure}[t]{0.3\textwidth}
        \centering
        \includegraphics[width=\textwidth]{Images/GT.png}
        \caption{Ground Truth}
        \label{fig:gt_k2}
    \end{subfigure}
    \hfill
    \begin{subfigure}[t]{0.3\textwidth}
        \centering
        \includegraphics[width=\textwidth]{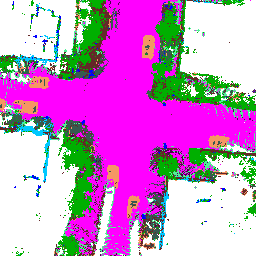}
        \caption{ConvBKI Zero-Shot}
        \label{fig:conv_k}
    \end{subfigure}
    \hfill
    \begin{subfigure}[t]{0.3\textwidth}
        \centering
        \includegraphics[width=\textwidth]{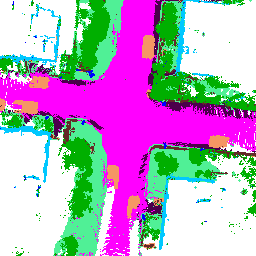}
        \caption{ConvBKI Re-trained Segmentation}
        \label{fig:conv_k_t}
    \end{subfigure}
    \begin{subfigure}[t]{0.3\textwidth}
        \centering
        \includegraphics[width=\textwidth]{Images/GT.png}
        \caption{Ground Truth}
        \label{fig:gt_k3}
    \end{subfigure}
    \hfill
    \hfill
    \begin{subfigure}[t]{0.3\textwidth}
        \centering
        \includegraphics[width=\textwidth]{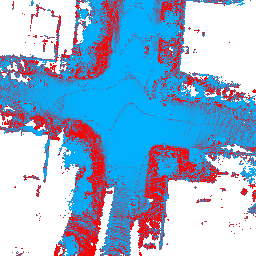}
        \caption{ConvBKI Variance Zero-Shot}
        \label{fig:conv_k_var}
    \end{subfigure}
    \hfill
    \begin{subfigure}[t]{0.3\textwidth}
        \centering
        \includegraphics[width=\textwidth]{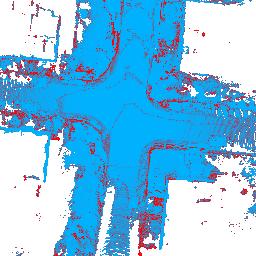}
        \caption{ConvBKI Variance Re-trained Segmentation}
        \label{fig:conv_k_t_var}
    \end{subfigure}
    \caption{Comparison of ConvBKI and Semantic MapNet on semantic mapping transferring from simulated to real data. Regions in white are unobserved by ConvBKI or in the ground truth. ConvBKI is generally successful, however suffers issues on ground segmentation due to LiDAR input. Ground segmentation is also a challenge on the simulated data, as the input semantic segmentation network is sometimes unable to distinguish sidewalk from road.} 
    \label{fig:KITTI_qualitative}
\vspace{-4mm}
\end{figure*}

\begin{table}[b]
% \vspace{2mm}
\vspace{-4mm}
% \footnotesize
\centering
\caption{Per-class comparison of ConvBKI and SemanticMapNet (SMNet) on simulated data. Key results are bolded. Performance of both algorithms is similar, except on the dynamic object categories. ConvBKI leaves traces which penalizes precision of dynamic objects and recall of ground surfaces.}
\resizebox{0.48\textwidth}{!}{
\begin{tabular}{l|l|cccccccccc}
% \toprule
% \multicolumn{3}{c|}{Method}
\multicolumn{1}{l}{\bf Method}&
\multicolumn{1}{l}{\bf Metric}&

\cellcolor{building}\rotatebox{90}{\color{white}Building} &
\cellcolor{barrier}\rotatebox{90}{\color{white}Barrier} &
\cellcolor{other}\rotatebox{90}{\color{white}Other} &
\cellcolor{pedestrian}\rotatebox{90}{\color{white}Pedestrian} &
\cellcolor{pole}\rotatebox{90}{\color{white}Pole} &
\cellcolor{road}\rotatebox{90}{\color{white}Road} &
\cellcolor{ground}\rotatebox{90}{\color{white}Ground} &
\cellcolor{sidewalk}\rotatebox{90}{\color{white}Sidewalk} & 
\cellcolor{vegetation}\rotatebox{90}{\color{white}Vege.} &
\cellcolor{vehicle}\rotatebox{90}{\color{white}Vehicle} \\ 

\hline 

\vspace{-2mm} \\
{\bf ConvBKI} & {\bf Prec.} & 59.8 & 1.3 & 51.5 & \textbf{10.5} & \textbf{29.3} & 85.1 & 5.2 & 91.2 & 57.4 & \textbf{21.3}\\
{\bf SMNet} & {\bf Prec.} & 67.7 & 1.3 & 43.5 & 31.5 & 50.2 & 87.2 & 6.06 & 92.6 & 60.3 & 56.5\\
\bottomrule
\vspace{-2mm} \\
{\bf ConvBKI} & {\bf Rec.} & 28.3 & 18.5 & 20.4 & \textbf{72.4} & 44.9 & \textbf{88.7} & 42.4 & \textbf{41.3} & 85.4 & 69.4\\
{\bf SMNet} & {\bf Rec.} & 34.2 & 13.2 & 26.0 & 35.4 & 55.3 & 98.1 & 22.5 & 67.7 & 88.9 & 53.0\\
\bottomrule
\vspace{-2mm} \\
{\bf ConvBKI} & {\bf IoU} & 23.8 & 1.2 & 17.1 & 10.1 & 21.5 & 76.8 & 4.9 & 39.7 & 52.3 & 19.4\\
{\bf SMNet} & {\bf IoU} & 29.4 & 1.2 & 19.4 & 20.0 & 35.7 & 85.8 & 5.0 & 64.2 & 56.1 & 37.7\\

\end{tabular}
}

\label{tab:precision_recall}
% \vspace{-4mm}
\end{table}

\textbf{Efficiency:} Key metrics on the test set of CarlaSC are summarized in Table \ref{tab:CarlaResults}. Compared to the learning-based baselines, ConvBKI builds and maintains a \textit{three dimensional} map using \textit{less memory} than the baselines take to operate in two dimensions. This difference is because the ConvBKI update is explicitly defined to achieve optimal efficiency by building off prior probabilistic mapping works. ConvBKI only requires two parameters for each semantic category to learn the geometry, for a total of 22 parameters. In contrast, the learning-based memory models contain nearly a hundred thousand times more parameters due to purely implicit operations. While implicit operations potentially enable more expressive functions to be learned, they are also less reliable as demonstrated in the next section. Additionally, ConvBKI achieves a quicker inference rate despite operating in 3-D due to maximally efficient operations. Note that the inference rate is different from Table \ref{tab:kitti_iou} due to the change in grid size.

\textbf{Training:} Another advantage of ConvBKI is the form of training data. ConvBKI trains on semantic segmentation ground-truth data which is more easily obtained than ground-truth BEV maps. Obtaining accurate ground-truth maps in the real world is a difficult challenge due to the presence of dynamic objects \cite{MotionSC}, requiring multiple viewpoints for accurate data. Semantic segmentation ground-truth data is more easily obtained, as evident by the vast number of on-road driving datasets. 

The primary trade-off of ConvBKI is highlighted by the BEV semantic segmentation performance. ConvBKI trains quickly in less than an epoch, but is gradually outperformed by implicit memory models as they train over the course of several days. With nearly a hundred thousand times more parameters, Semantic MapNet is able to outperform ConvBKI on BEV segmentation. ConvBKI obtains a higher mIoU of 26.7$\%$ after one epoch of training compared to the implicit memory models, but is surpassed by the GRU after several days of training which obtains an mIoU of 35.4$\%$.

\textbf{Dynamic Objects:} {\color{black}The lower mIoU of ConvBKI can be explained by examining the precision and recall of each semantic category in Table \ref{tab:precision_recall}. ConvBKI is comparable to Semantic MapNet across most semantic categories, however is penalized on the dynamic classes due to the lack of a dynamic object propagation step considering object motion}. ConvBKI recurrently incorporates semantically segmented points into the semantic map without separate treatment of dynamic objects, leaving traces. Measured on BEV mapping, traces cause lower precision of the pedestrian and vehicle classes, as well as lower recall of the sidewalk and road classes. Dynamic methods have already been proposed which fit within the BKI framework \cite{DynamicBKI}, however they do not fit into the scope of this paper and we have left the integration of dynamic mapping with our accelerated network as future work.

A confusion matrix of ConvBKI is shown in Fig. \ref{fig:confusion_matrix}, and a table of the precision, recall, and IoU per class compared to the fully trained Semantic MapNet is shown in Table \ref{tab:precision_recall}. In general, the performance between the two methods is very similar amongst all classes, with a few exceptions which {\color{black}motivates future work. }
Some notable columns include pedestrian and vehicle, where ConvBKI has a higher recall yet much lower precision due to artifacts from dynamic objects. In ConvBKI, dynamic objects are treated the same as static objects, which leave traces in the map. When projecting from 3D to 2D, the traces cause ground tiles to be incorrectly labeled as a dynamic object, thereby reducing precision. For the pedestrian class, many pedestrian labels are incorrectly applied to sidewalk tiles, and the same for vehicle predictions to the road class. Note that this is due to the highly dynamic nature of the CarlaSC simulated dataset, and would not be present in static scenes where performance would be higher. The same effect is seen on the road and sidewalk classes, where ConvBKI has a similar precision but lower recall since the BEV projection captures traces left behind by dynamic objects. 

\textbf{Reliability:} Another key result is highlighted by the groups of ground classes and building/barrier classes. Our method is ultimately reliant on the input semantic predictions of the semantic segmentation network. Although ConvBKI may smooth out noise, it is incapable of resolving errors where the segmentation predicts temporally consistent incorrect labels. However, the predictable nature of our algorithm may also be seen as an advantage for improved reliability, as observed occupied voxels will always be modeled within our map, even if the label is wrong. 

In Fig. \ref{fig:CarlaSC_qualitative}, a pedestrian walks along the bottom right on a sidewalk. ConvBKI leaves a trace as it does not yet handle dynamic objects, while Semantic MapNet completely removes the pedestrian. This occurs because deep neural networks are constructed to optimize a specific metric, in this case mIoU, and may hallucinate to create the best performance at the cost of wrong predictions. In contrast, ConvBKI is heavily penalized on the pedestrian and road class for leaving traces. In safety-critical environments such as self-driving cars, however, it is vital to have a trustworthy system which does not collide with pedestrians or other difficult semantic categories, even if the semantic label is incorrect. 

\subsubsection{Transfer to Real Data}
Next, we highlight the reliability of our network by transferring from simulated CarlaSC to real-world Semantic KITTI. The goal of this section is to evaluate the uncertainty quantification ability of ConvBKI under stress, as well as the ability of ConvBKI and Semantic MapNet to translate to a different dataset. Due to the limited parameters and explicit domain knowledge embedded within ConvBKI, it is more reliable than purely implicit memory models and capable of more successfully bridging the sim-to-real gap. Additionally, the explicit operations of ConvBKI allows the semantic segmentation network to be directly swapped for one trained on the specific environment. 

First, we compare ConvBKI with Semantic MapNet on the validation set of Semantic KITTI without any re-training of the semantic segmentation network or memory network. Next, we re-train the semantic segmentation network without modifying the memory networks to demonstrate the advantage of the explicit domain knowledge of ConvBKI.

\begin{table}[b]
% \vspace{2mm}
\vspace{-4mm}
% \footnotesize
\centering
\caption{Quantitative results transferring sim-to-real from CarlaSC \cite{MotionSC} to validation sequence of Semantic KITTI \cite{SemanticKITTI}. Results for each method are measured once directly from sim-to-real (direct) and once with the semantic segmentation network replaced, indicated by (tran.).}
\resizebox{0.48\textwidth}{!}{
\begin{tabular}{l|cccccccccc|c}
% \toprule
% \multicolumn{3}{c|}{Method}
\multicolumn{1}{l}{\bf Method}&

\cellcolor{building}\rotatebox{90}{\color{white}Building} &
\cellcolor{pedestrian}\rotatebox{90}{\color{white}Pedestrian} &
\cellcolor{pole}\rotatebox{90}{\color{white}Pole} &
\cellcolor{road}\rotatebox{90}{\color{white}Road} &
\cellcolor{ground}\rotatebox{90}{\color{white}Ground} &
\cellcolor{sidewalk}\rotatebox{90}{\color{white}Sidewalk} & 
\cellcolor{vegetation}\rotatebox{90}{\color{white}Vege.} &
\cellcolor{vehicle}\rotatebox{90}{\color{white}Vehicle} &
\rotatebox{90}{\bf mIoU (\%)}\\ 

\hline 

\vspace{-2mm} \\
{\bf ConvBKI (direct)} & 21.1 & 17.4 & 2.8 & 41.8 & 19.4 & 11.4 & 49.3 & 30.1 & 24.2\\
\bottomrule

\vspace{-2mm} \\
{\bf SMNet (direct)} & 22.2 & 3.7 & 7.0 & 40.3 & 22.2 & 18.4 & 49.7 & 27.8 & 23.9\\
\bottomrule

\vspace{-2mm} \\
{\bf ConvBKI (tran.)} & {\bf 58.5} & {\bf 23.6} & {\bf 12.3} & {\bf 66.7} & {\bf 64.7} & {\bf 40.5} & {\bf 61.6} & {\bf 45.1} & {\bf 46.6}\\
\bottomrule

\vspace{-2mm} \\
{\bf SMNet (tran.)} & 0.0 & 0.0 & 0.0 & 18.5 & 13.7 & 1.3 & 3.2 & 0.3 & 4.6\\

\end{tabular}
}

\label{tab:kitti_transfer}
\end{table}

Ground-truth data is obtained from the Semantic KITTI Semantic Scene Completion task by re-constructing voxels over the same bounds as CarlaSC. Normally, the Semantic Scene Completion task excludes the scene posterior to the ego vehicle, so we re-run the Semantic KITTI voxelizer to ensure BEV map bounds match between CarlaSC and Semantic KITTI voxels. Next, we obtain BEV ground truth from the voxelized scenes following the same procedure as before. ConvBKI and Semantic MapNet are then run directly on Semantic KITTI without any re-training to attempt zero-shot domain transfer from simulation to the real world. 

\textbf{Direct Transfer:} ConvBKI is able to more successfully bridge the sim-to-real gap due to explicit domain knowledge which is less susceptible to noise than complex implicit operations. However, error is still introduced by the semantic segmentation network SPVCNN which has not been re-trained. The most common error in ConvBKI is incorrectly labeling other types of ground as road, which is a common systematic error of LiDAR-based semantic segmentation.

Quantitative findings are summarized in Table \ref{tab:kitti_transfer}, and predicted BEV maps are shown in Fig. \ref{fig:KITTI_qualitative}. The performance of Semantic MapNet decreases from a mIoU of 35.4$\%$ to 23.9$\%$ when bridging the sim-to-real gap, whereas ConvBKI has a less significant decrease from 26.7$\%$ on simulated data to 24.2$\%$ when directly transferring from simulated data to real data. On the new data-set, ConvBKI is able to outperform Semantic MapNet due to a less significant decline in performance. Additionally, when bridging the sim-to-real gap, Semantic MapNet becomes unable to effectively recognize the pedestrian class. As previously stated in Fig. \ref{fig:CarlaSC_qualitative}, deep learning methods for mapping learn to hallucinate the scene and may unreliably predict the absence of critical objects such as pedestrians to optimize a metric. When transferring from simulation to the real world, pedestrians become significantly more difficult to identify, causing the network to omit pedestrians. In contrast, our method still predicts the presence of pedestrians but is penalized due to leaving artifacts from dynamic objects as previously discussed. 

\textbf{Explicit Knowledge:} Another prominent result from Table \ref{tab:kitti_transfer} is the difference in performance between ConvBKI and Semantic MapNet when swapping the semantic segmentation network. Since ConvBKI learns explicit distributions over semantic categories, the kernels directly transfer to other semantic segmentation network inputs. The semantic segmentation network may be easily replaced with different weights, in this case, trained on Semantic KITTI. ConvBKI can directly transfer to a new semantic segmentation network since the semantic classes remain the same and may be directly mapped. In contrast, semantic classes in the latent space have no guarantee to correlate.

As expected, the results of ConvBKI significantly improve when the semantic segmentation network performance is improved. Swapping the semantic segmentation network causes the mIoU of ConvBKI to increase from 24.2$\%$ to 46.6$\%$. However, since there is no guarantee that features will match in the latent space, the performance of Semantic MapNet decreases when the segmentation network is replaced to 4.6$\%$. These results demonstrate that the world knowledge learned by ConvBKI can readily transfer to other datasets and may be easily improved through the substitution of the semantic segmentation network. 

\textbf{Variance:} ConvBI also has the ability to quantify uncertainty, which is highlighted in Fig. \ref{fig:KITTI_qualitative}. Transferring to a new dataset is challenging for the semantic segmentation network, causing high variance in the variance map of ConvBKI. As the segmentation network is swapped, causing an improvement in semantic segmentation, the level of variance in the map decreases. This quality demonstrates the utility of uncertainty quantification, as the variance map provides a measure of confidence for important downstream tasks such as traversability used in path planning. When exposed to new data, the uncertainty can provide warnings to proceed cautiously or stop operation in order to avoid costly errors. 

% \begin{figure}[t]
%     \centering
%     \begin{subfigure}[t]{0.4\textwidth}
%         \centering
%         \includegraphics[width=\textwidth]{Images/bki_net_nofilter.png}
%         %\caption{Single}
%         \label{fig:bki_raw}
%     \end{subfigure}
%     \\[-2ex]
%     \begin{subfigure}[t]{0.4\textwidth}
%         \centering
%         \includegraphics[width=\textwidth]{Images/bki_net_variance_map.png}
%         %\caption{Per Class}
%         \label{fig:bki_variance}
%     \end{subfigure}
%     \\[-2ex]
%     \begin{subfigure}[t]{0.4\textwidth}
%         \centering
%         \includegraphics[width=\textwidth]{Images/bki_net_filtered.png}
%         %\caption{Compound}
%         \label{fig:bki_filtered}
%     \end{subfigure}
%     \caption{Example map produced by ConvBKI Compound on the validation set of Semantic KITTI. The expected semantic map is shown in the top image, and the variance is shown in the middle, where red indicates high variance and blue indicates low variance. Voxels with high variance or uncertainty (e.g., $\mathcal{V}[\alpha^c_*]$ > 0.01) correlate with incorrect semantic labels. Additionally, variance provides insight into which regions of the map are unexplored and may be hazardous.
%     } 
%     \label{fig:SemanticMap}
% \label{fig:bki_map}
% \vspace{-4mm}
% \end{figure}
 
\subsection{Ablation Studies}\label{Sec:Ablation}
For our final quantitative results, we analyze design choices on the Semantic KITTI dataset. For this study, we re-train ConvBKI on the validation set of Semantic KITTI for a single epoch and evaluate semantic segmentation on the same set in order to study the upper bound of the expected performance gain. We train and test on a voxel grid with bounds of [-20, -20, -2.6] to [20, 20, 0.6] $\m$ along the (X, Y, Z) axes, where points outside of the voxel grid are discarded and not measured in the results. Frequencies are again calculated by the average update latency, with additional results on computation time for input discretization and free space sampling. 

\begin{figure}[t]
    \centering
    
    \begin{subfigure}[t]{0.15\textwidth}
        \centering
        \includegraphics[width=\textwidth]{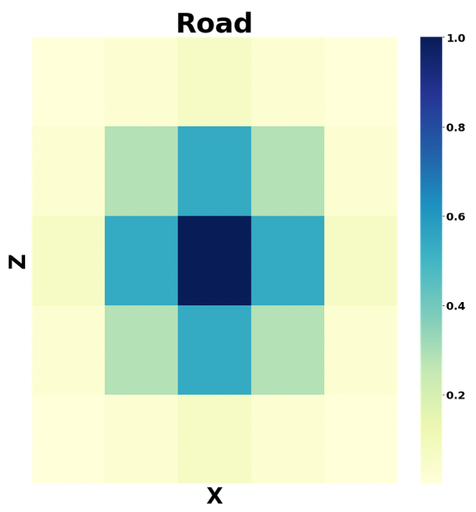}
        %\caption{Single}
        \label{fig:sing_road}
    \end{subfigure}
    \hfill
    \begin{subfigure}[t]{0.15\textwidth}
        \centering
        \includegraphics[width=\textwidth]{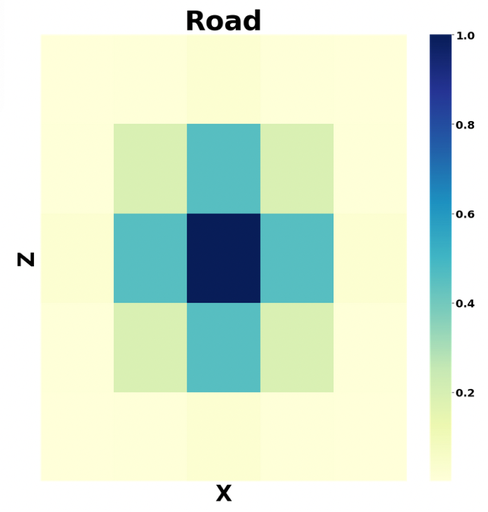}
        %\caption{Per Class}
        \label{fig:pc_road}
    \end{subfigure}
    \hfill
    \begin{subfigure}[t]{0.15\textwidth}
        \centering
        \includegraphics[width=\textwidth]{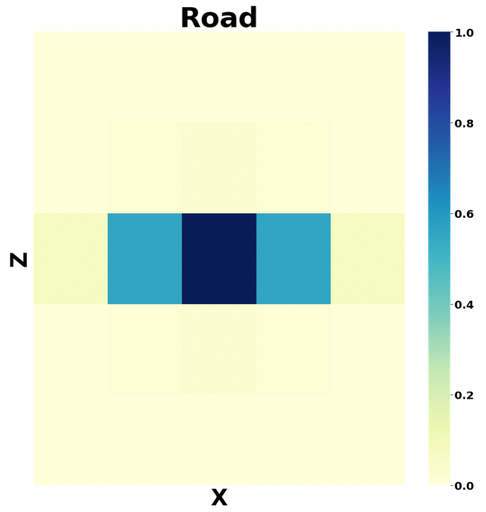}
        %\caption{Compound}
        \label{fig:comp_road}
    \end{subfigure}
    \hfill

    \begin{subfigure}[b]{0.15\textwidth}
        \centering
        \includegraphics[width=\textwidth]{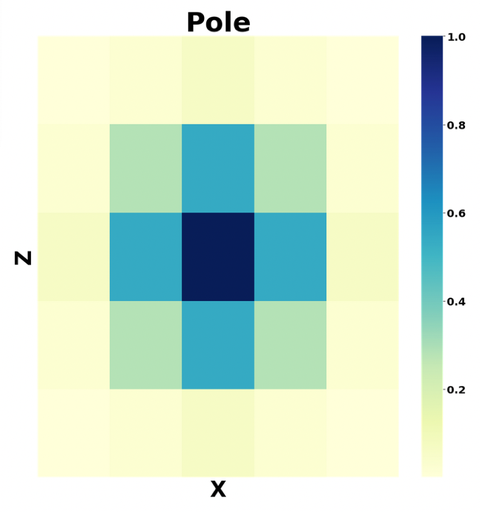}
        \caption{Single}
        \label{fig:sing_pole}
    \end{subfigure}
    \hfill
    \begin{subfigure}[b]{0.15\textwidth}
        \centering
        \includegraphics[width=\textwidth]{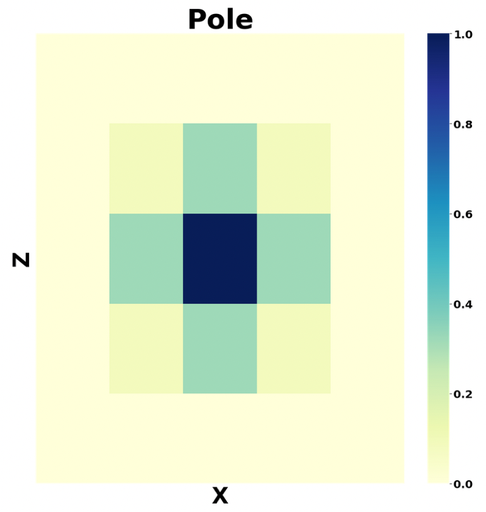}
        \caption{Per Class}
        \label{fig:pc_pole}
    \end{subfigure}
    \hfill
    \begin{subfigure}[b]{0.15\textwidth}
        \centering
        \includegraphics[width=\textwidth]{Images/CompoundPole.png}
        \caption{Compound}
        \label{fig:comp_pole}
    \end{subfigure}
    \caption{Illustration of kernels learned by ConvBKI on the road and pole semantic classes, plotted at $\Delta Y=0$. Adding degrees of freedom increases expressivity by allowing the kernel to learn class-specific geometry.} 
    \label{fig:KernelDemo}
\vspace{-4mm}
\end{figure}

\textbf{Resolution:} First, we study the effect of voxel resolution on the inference time and mIoU of ConvBKI. We compare ConvBKI with resolutions 0.1, 0.2 and 0.4~$\m$, and a constant filter size $f=5$ for all models. Table~\ref{tab:resolution} indicates that a finer resolution can increase performance; however, the segmentation difference between 0.2 and 0.1~$\m$ resolution is marginal at the cost of greater memory and slower inference. For real-time driving applications, this suggests that 0.2~$\m$ resolution may be a strong middle ground. Note that the optimal resolution will vary between applications, such as finer resolution for indoor mobile robots or with LiDAR resolution. 
 
 \textbf{Filter Size:} Next, we study the effect of the filter size on inference time and performance summarized in Table~\ref{tab:Filter_Size}. While a larger filter size increases the receptive field of the kernel and potentially improves the predictive capability as a result, filter size also cubically increases computation cost. Therefore, identifying a balance between filter size and computational efficiency is important for real-time applications. We study filters of size $f=$ 3, 5, 7, and 9 at a resolution of 0.2~$\m$. Table \ref{tab:Filter_Size} demonstrates that filter sizes can improve segmentation accuracy, however, quickly increase run-time. In practice, a filter size of 5 or 7 may be optimal, as a filter size of 9 offers little improvement with a large increase in computational cost. As with resolution, optimal filter size will vary between applications and LiDAR resolution since there will always be a trade-off between performance and latency.

 % \Parker{Could you make Tables IV and V into graphs? For Table 5 you could represent latency as a line with 1 standard dev shown and mIoU as a bar. I find graphs look nicer than tables, but thats just my opinion.}

 \begin{table}[b]
 \vspace{-4mm}
\centering
\caption{Ablation study of voxel resolution on Semantic KITTI sequence 8 for compound ConvBKI with filter size $f=5$.}
\scriptsize
\begin{tabular}{|c|c|c|c|c|}
 \toprule
 \textbf{Resolution} & \textbf{mIoU (\%)} & \textbf{Frequency ($\mathrm{Hz}$)} & \textbf{Mem. (GB)} \\
 \bottomrule
 \hline
 N/A (Input) & 54.6 & n/a & n/a\\
 \hline
 0.4~$\m$ & 58.2 & \textbf{301.4} & \textbf{2.4}\\
 \hline
 0.2~$\m$ & \textbf{59.3} & 216.8 & 2.7\\
 \hline
 0.1~$\m$ & 59.0 & 65.3 & 5.0\\
 \bottomrule
 \end{tabular}
\label{tab:resolution}
\end{table}

\begin{table}[b]
\vspace{-4mm}
\centering
\caption{Ablation study of filter size on Semantic KITTI sequence 8 for compound ConvBKI with resolution $0.2 \m$.}
\scriptsize
\begin{tabular}{|c|c|c|c|}
 \toprule
 \textbf{Filter Size} & \textbf{mIoU (\%)} & \textbf{Frequency ($\mathrm{Hz}$)}\\
 \bottomrule
 \hline
 N/A (Input) & 54.6 & n/a\\
 \hline
 $f=3$ & 59.0 & \textbf{260.2}\\
 \hline
 $f=5$ & 59.3 & 216.8\\
 \hline
 $f=7$ & 59.5 & 161.5\\
 \hline
 $f=9$ & \textbf{59.6} & 117.0\\
 \bottomrule
 \end{tabular}
\label{tab:Filter_Size}
\end{table}

\textbf{Expressivity:} We illustrate the kernels produced by our map to emphasize the underlying operations that match expectations of semantic-geometric distributions, such as that poles should be tall and roads should be flat. Fig.~\ref{fig:KernelDemo} demonstrates the kernels learned by variations of the ConvBKI layer for single, per class, and compound kernels. Each variation of \mbox{ConvBKI} improves potential semantic-geometric expressiveness. \mbox{ConvBKI} Single learns one semantic-geometric distribution shared between all classes. However, semantic classes do not share the same geometry in the real world. ConvBKI Per Class adds the capability to learn a unique distribution for each semantic category but is still restricted geometrically. ConvBKI Compound learns a more complex geometric distribution, which can be more expressive for classes with specific shapes.

\begin{figure}[t]
    \includegraphics[width=0.9\linewidth]{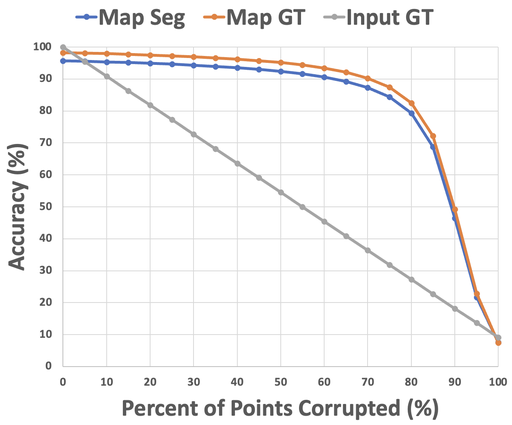}
    \centering
    \caption{Ablation study on the smoothing effect of BKI in the presence of noisy segmentation predictions {\color{black}due to aleatoric uncertainty. The horizontal axis indicates the portion of semantic segmentation labels replaced with a random choice from the set of semantic classes. Results are computed as the semantic segmentation accuracy of ConvBKI over every point cloud of Semantic KITTI sequence 8. Two optimal baselines are included for comparison. Input GT refers to the ground truth semantic segmentation without ConvBKI. Map GT refers to the predictions of ConvBKI with ground truth semantic segmentation labels as input. Map Seg refers to the predictions of ConvBKI with neural network predictions as input. Most notably, the segmentation smoothing of ConvBKI minimizes the effect of white noise since spurious predictions are effectively averaged out.} 
    }
    \label{fig:NoiseVsAccuracy}
    \vspace{-4mm}
\end{figure}

\begin{figure}[t]
    \includegraphics[width=0.9\linewidth]{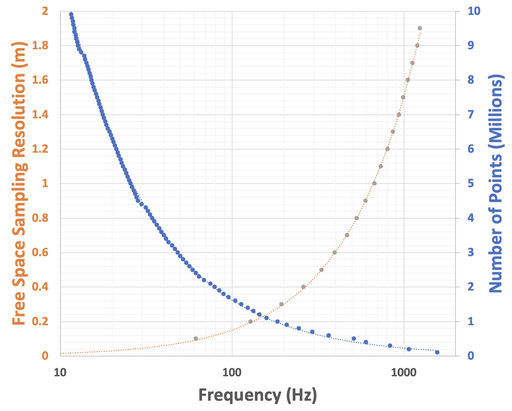}
    \centering
    \caption{{\color{black}Frequency of sampling free space points compared to the free space resolution in meters on GPU, and frequency of point insertion into the convolutional input grid compared to the number of points. As free space sampling resolution becomes more fine, the frequency of free space sampling decreases and the number of points grows. Additionally, as the number of points increases, the frequency of input grid construction decreases. While the computation cost of constructing the input grid is dependent on the number of points and free space resolution, the computation complexity of the ConvBKI operation remains constant and is instead dependent on the filter size and map size.}
    }
    \label{fig:FreeSpaceFrequency}
    % \vspace{-4mm}
\end{figure}

\textbf{Robustness:} {\color{black}We evaluate the robustness of ConvBKI from aleatoric uncertainty in Fig. \ref{fig:NoiseVsAccuracy} compared to optimal baselines. Fig. \ref{fig:NoiseVsAccuracy} demonstrates a plot of the percentage of noisy semantic segmentation input labels to ConvBKI versus the overall accuracy. As an optimal baseline, we include a plot of the segmentation accuracy of ground truth labels without mapping (Input GT). As expected, the plot of Input GT begins at an accuracy of $100 \%$ and decreases linearly as the percentage of noisy labels increases. Next, we examine the performance of ConvBKI with the ground truth semantic segmentation labels as input (Map GT). Due to compression of points into voxels, the accuracy of Map GT without noise is slightly less than Input GT, but quickly surpasses the performance of Input GT as the amount of noise increases. Next, we plot ConvBKI with predictions from the semantic segmentation neural network as input, denoted Map Seg. ConvBKI performance with semantic segmentation predictions is similar to the optimal baseline with ground truth segmentation, but the gap between accuracies decreases as the amount of noise increases. 

The spatial smoothing operation of ConvBKI reduces the effects of noise at low levels. At high levels of noise, the update step is dominated by noise, causing a sharp decrease in performance.
As the amount of noisy predictions increases, the variance of the maximum likelihood prediction also increases due to Eq. \eqref{eq:Variance}.

Robustness to epistemic or systemic uncertainty is more difficult for ConvBKI to capture, and is only encapsulated in the variance calculation if objects are more easily distinguishable from different viewpoints. For example, in Fig. \ref{fig:KITTI_qualitative}, road and sidewalk are systematically challenging to distinguish between from LiDAR, but are easier to identify depending on the viewpoint, causing the high variance about misclassified sidewalk regions in ConvBKI zero-shot.}

\textbf{Free Space Sampling:} {\color{black}Free space sampling is important for planning frameworks, and is a computationally expensive component of mapping. We sample free space points at fixed intervals along each ray which increases the total number of points. As the number of points increases, the computational complexity required to insert points into the observation grid increases, however the update complexity of the convolutional filter remains constant. 

Fig. \ref{fig:FreeSpaceFrequency} demonstrates a plot of the frequency of sampling free space points for a point cloud of $120,000$ points dependent on the free space sampling resolution between $0$ and $2 \m$. Since the free space sampling equation is vectorized and computed on GPU, the frequency is still well above $10$ Hz with a resolution finer than $0.1 \m$. As the number of points increases due to free space sampling or large data such as fine resolution images, so does the cost of constructing the input for the convolution layer. Fig. \ref{fig:FreeSpaceFrequency} also demonstrates the cost of input grid construction compared to the number of input points. Even at 10 million points, grid insertion is still greater than $10$ Hz. Consideration of free space sampling resolution and size of input is necessary for real-time application, as the number of points can rapidly grow depending on parameter choices. 
}

\begin{figure}[t]
    \includegraphics[width=0.9\linewidth]{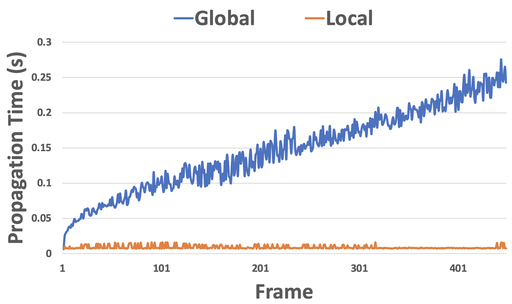}
    \centering
    \caption{Propagation time of global and local mapping on Semantic KITTI sequence 8. Since ConvBKI operates on a dense grid, querying the local grid quickly becomes the bottleneck. Local mapping enables accelerated by maintaining the dense local map in memory.
    }
    \label{fig:LocalVsGlobal}
    \vspace{-4mm}
\end{figure}

\textbf{Local Mapping:} Last, we study our proposed local mapping method, which we employ for quicker inference. While the update step of ConvBKI may run consistently at high inference rates, the bottleneck of global mapping is querying the dense local region to perform the depthwise separable convolution update. As discussed in Section \ref{sec:Local Mapping}, we constrain memory to only maintain the dense local region so the bottleneck may be removed. A comparison of the latency of ConvBKI with global mapping and local mapping is shown in Fig. \ref{fig:LocalVsGlobal} on the propagation step. Over time, the size of the global map increases, slowing the dense query operation linearly with the number of occupied voxels. In contrast, the local mapping variant has a constant propagation time and removes the bottleneck by restricting map memory to a fixed-size grid which ConvBKI naturally operates on.

\begin{figure}[b]
    \includegraphics[width=\linewidth]{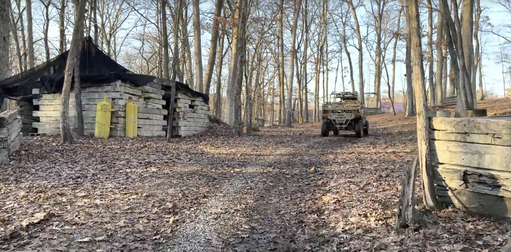}
    \centering
    \caption{Image from off-road testing. Our dataset includes many semantic categories which are not encountered in RELLIS-3D \cite{Rellis3D}, in order to test the reliability of our algorithm in difficult situations.
    }
    \label{fig:mrzr}
    % \vspace{-4mm}
\end{figure}

\begin{figure}[t]
    \centering
    \begin{subfigure}[t]{0.23\textwidth}
        \centering
        \includegraphics[width=\textwidth]{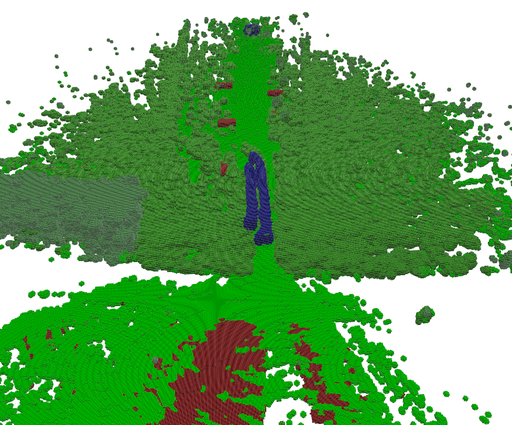}
        \caption{Semantic Predictions}
        \label{fig:rellis_sem}
    \end{subfigure}
    \hfill
    \begin{subfigure}[t]{0.23\textwidth}
        \centering
        \includegraphics[width=\textwidth]{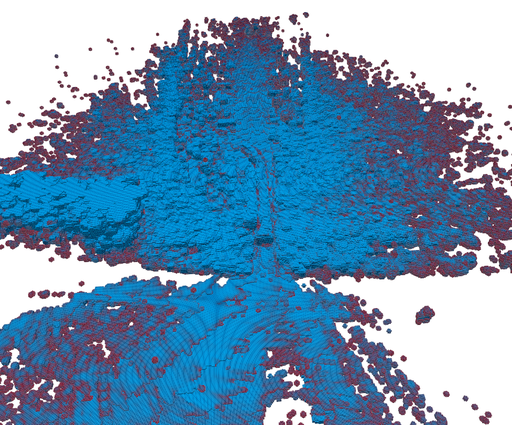}
        \caption{Variance}
        \label{fig:rellis_var}
    \end{subfigure}
    \caption{Expectation and variance of the semantic map on RELLIS-3D dataset. The data set is the largest open-source LiDAR semantic segmentation dataset available at the time of writing but is limited in the number of classes. Typical semantic categories encountered include rubble in red, pedestrian in blue, bush in dark green, and grass in light green, in generally easy scenarios. In contrast, our dataset includes more semantic categories unseen by the robot during training as well as more difficult perceptual conditions.}
    \label{fig:rellis}
\end{figure}

\section{Off-road}\label{sec:offroad}
Finally, we combine ConvBKI with a semantic segmentation LiDAR network into a Robot Operating System (ROS) package for end-to-end testing. We gather a new off-road data set to study the reliability of our system under perceptually challenging scenarios, including sparse, unstructured data, and when encountering semantic categories previously unseen during training. Although these conditions produce errors, ConvBKI produces quantifiable uncertainty per voxel, which can provide valuable information on when the robot is leaving its operating domain in order to prevent unforeseen errors. An example frame from our data set is shown in Fig. \ref{fig:mrzr}, which illustrates the test vehicle as well as the typical scenes encountered in our data set. 

The ROS package is available publicly at \href{https://github.com/UMich-CURLY/BKI_ROS}{https://github.com/UMich-CURLY/BKI_ROS}. We pre-process poses using an off-the-shelf LiDAR-based SLAM algorithm \cite{LIOSAM}, and perform real-time mapping from recorded ROS bags of real-world scenes. Semantic maps and variance heat maps are published to ROS for visualization as voxel grids. For our semantic segmentation network, we again use SPVCNN \cite{SPVNAS} for real-time LiDAR semantic segmentation. 

\subsection{Training}
We train ConvBKI and SPVCNN on the popular off-road driving data set RELLIS-3D \cite{Rellis3D}. RELLIS-3D contains per-point LiDAR semantic segmentation ground truth labels in a format and size similar to Semantic KITTI \cite{SemanticKITTI}. The LiDAR data set contains several imbalanced semantic categories, including grass, tree, and bush, as well as concrete, mud, person, puddle, rubble, barrier, log, fence, and vehicle. The three most common categories are grass, tree, and bush. Scenes in RELLIS-3D are generally well structured, consisting of a robot following a well-defined path. An example semantic and variance map from RELLIS-3D is shown in Fig. \ref{fig:rellis}. On RELLIS-3D, the variance is low except for some regions at the parameters of the map, which are sparsely sampled. Data is captured from a 64-channel LiDAR for high-definition point clouds. 

Transferring from RELLIS-3D is particularly challenging as the environments and semantic classes encountered are very different, despite both being off-road data sets. As seen in Fig. \ref{fig:mrzr}, there is no clearly defined path in our data set, and a multitude of new classes are encountered. The image displays our test vehicle driving by large trees on an obscure path covered by trees, with buildings to either side. In RELLIS-3D, there is no building class, and categories such as log, fence, or barrier are rarely seen. These class imbalances and anomalies create unique conditions to test the uncertainty quantification of ConvBKI under stress. 

\subsection{Testing}
\begin{figure}[t]
    \includegraphics[width=0.99\columnwidth]{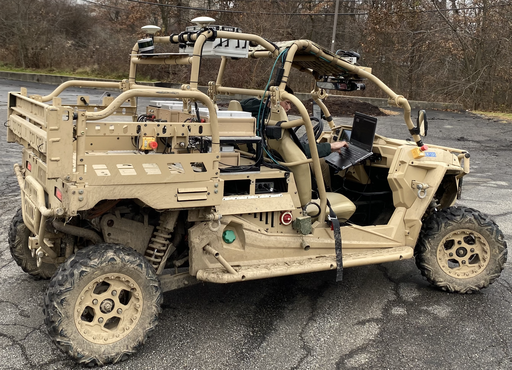}
    \centering
    \caption{Off-road test vehicle equipped with a full sensor suite and ROS. 
    }
    \label{fig:close_car}
    % \vspace{-4mm}
\end{figure}

Our data set is gathered with the support of Neya Systems from their full-size military test vehicle, the Polaris MRZR, equipped with a perception sensor suite and Robot Operating System (ROS). The sensor suite includes a 64-channel front-mounted LiDAR, 16-channel rear-mounted LiDAR, stereo camera, IMU, high-precision gyroscope, and GPS. An image of the test vehicle is shown in Fig. \ref{fig:close_car}. Since localization is out of the scope of this paper, we pre-process data using LIO-SAM \cite{LIOSAM}. LIO-SAM registers point clouds from point cloud and IMU data, providing pose transformations of the LiDAR sensor at each keyframe. After removing points from the front LiDAR within a 1.5 $\m$ radius around the LiDAR to avoid points from the ego-vehicle, there are approximately $40$ thousand points per point cloud, sampled at $10$ Hz. SLAM in off-road driving scenarios is particularly challenging, leaving a small amount of noise in the pose. This noise creates a blurring effect of dense foliage, such as the bushes in Fig. \ref{fig:rellis}. 

\begin{figure*}[t]
    \centering
    \begin{subfigure}[t]{0.49\textwidth}
        \centering
        \includegraphics[width=\textwidth]{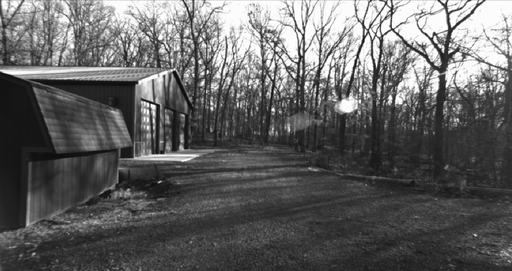}
        \label{fig:neya_cam_hill}
    \end{subfigure}
    \hfill
    \begin{subfigure}[t]{0.49\textwidth}
        \centering
        \includegraphics[width=\textwidth]{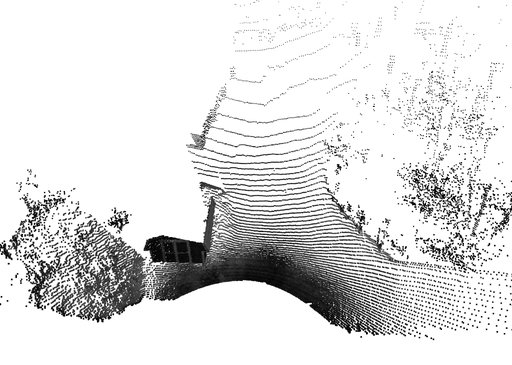}
        \label{fig:neya_pc_hill}
    \end{subfigure}  
    \hfill
    \begin{subfigure}[t]{0.49\textwidth}
        \centering
        \includegraphics[width=\textwidth]{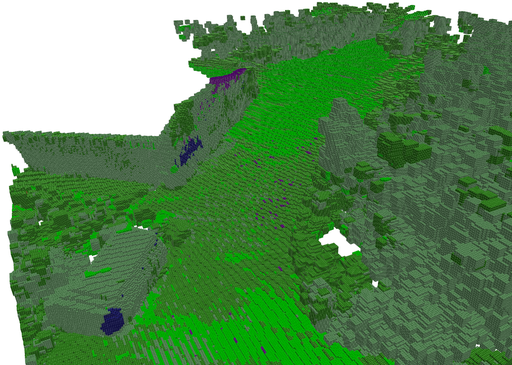}
        \label{fig:neya_hill_semantic}
    \end{subfigure}  
    \begin{subfigure}[t]{0.49\textwidth}
        \centering
        \includegraphics[width=\textwidth]{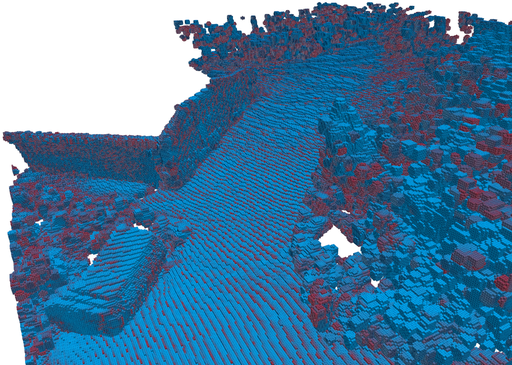}
        \label{fig:neya_hill_variance}
    \end{subfigure}
    \caption{Example raw data and output from off-road testing. The vehicle drives by two buildings on the left and a collection of trees to the left and front. Trees are in a light brown color, bushes in a dark green, grass in a light green, and pedestrian or pole in dark blue. The small patches of purple in the path are classified as rubble. Since buildings were not encountered during training, points are classified by the closest semantic category the network has seen. Taller portions resemble trees or pedestrians, while flatter objects resemble ground categories such as grass or bush. The point cloud is sparse around clustered bushes and trees, which causes higher variance in dense foliage. Other areas with high variance include misclassified bush and grass on the larger building, as well as the foliage at the end of the path, which has sparsely been observed by the robot.}
    \label{fig:neya_data1}
\end{figure*}

\begin{figure*}
    \centering
    \begin{subfigure}[t]{0.24\textwidth}
        \centering
        \includegraphics[width=\textwidth]{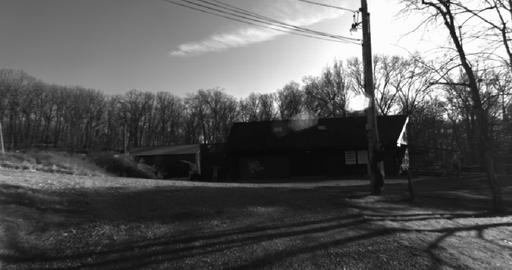}
        \label{fig:neya_cam_building}
    \end{subfigure}
    \hfill
    \begin{subfigure}[t]{0.24\textwidth}
        \centering
        \includegraphics[width=\textwidth]{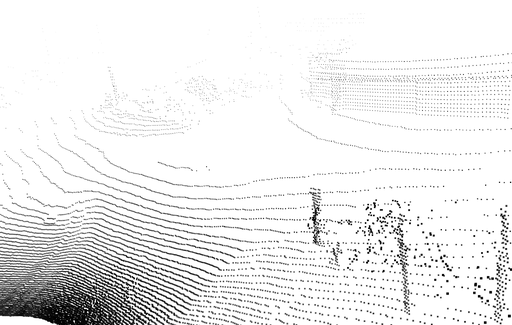}
        \label{fig:neya_pc_building}
    \end{subfigure}  
    \hfill
    \begin{subfigure}[t]{0.24\textwidth}
        \centering
        \includegraphics[width=\textwidth]{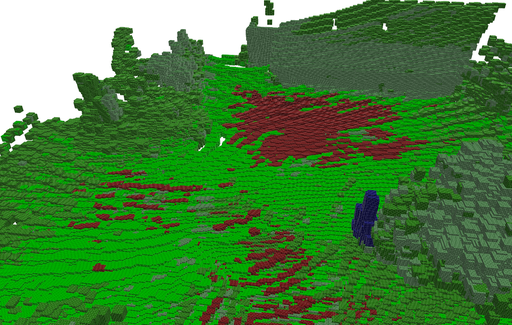}
        \label{fig:neya_building_semantic}
    \end{subfigure}  
    \hfill
    \begin{subfigure}[t]{0.24\textwidth}
        \centering
        \includegraphics[width=\textwidth]{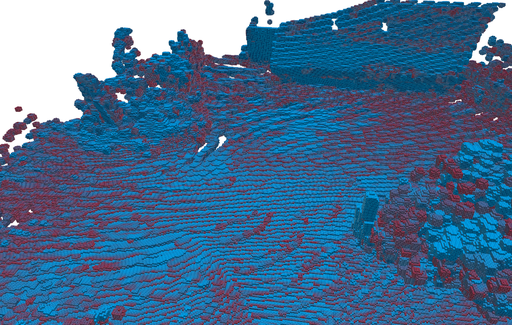}
        \label{fig:neya_building_variance}
    \end{subfigure}
    \hfill
    \begin{subfigure}[t]{0.24\textwidth}
        \centering
        \includegraphics[width=\textwidth]{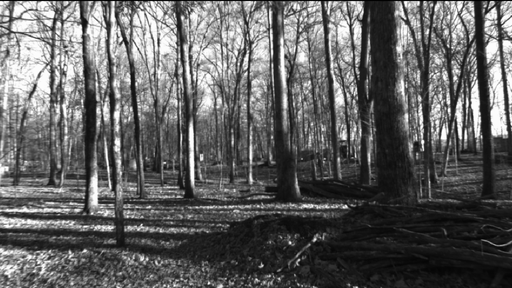}
        \label{fig:neya_cam_trees}
    \end{subfigure}
    \hfill
    \begin{subfigure}[t]{0.24\textwidth}
        \centering
        \includegraphics[width=\textwidth]{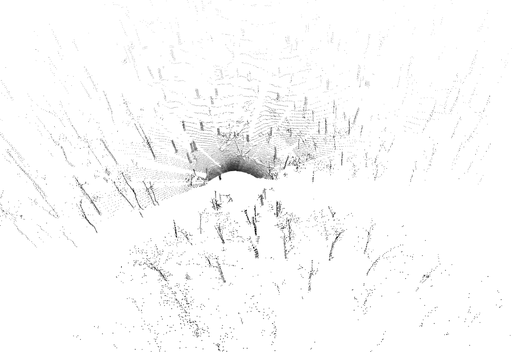}
        \label{fig:neya_pc_trees}
    \end{subfigure}  
    \hfill
    \begin{subfigure}[t]{0.24\textwidth}
        \centering
        \includegraphics[width=\textwidth]{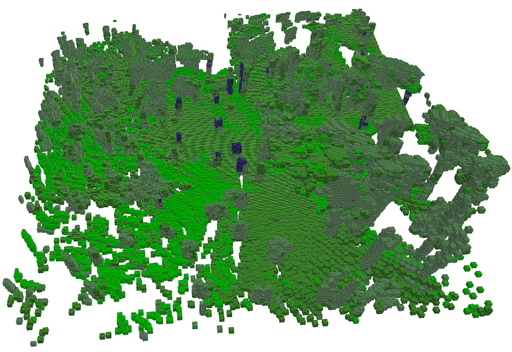}
        \label{fig:neya_trees_semantic}
    \end{subfigure}  
    \begin{subfigure}[t]{0.24\textwidth}
        \centering
        \includegraphics[width=\textwidth]{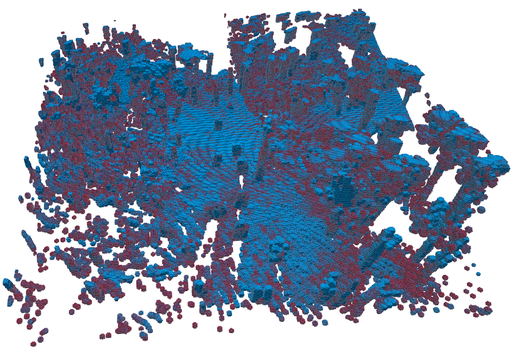}
        \label{fig:neya_trees_variance}
    \end{subfigure}
    \hfill
    \begin{subfigure}[t]{0.24\textwidth}
        \centering
        \includegraphics[width=\textwidth]{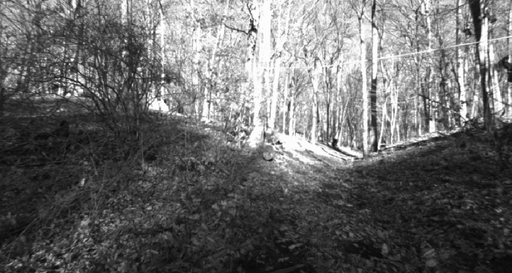}
        \label{fig:neya_cam_trees2}
    \end{subfigure}
    \hfill
    \begin{subfigure}[t]{0.24\textwidth}
        \centering
        \includegraphics[width=\textwidth]{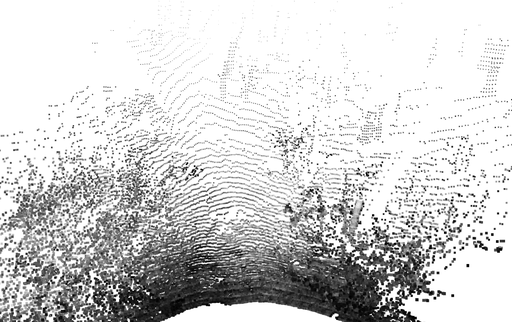}
        \label{fig:neya_pc_trees2}
    \end{subfigure}  
    \hfill
    \begin{subfigure}[t]{0.24\textwidth}
        \centering
        \includegraphics[width=\textwidth]{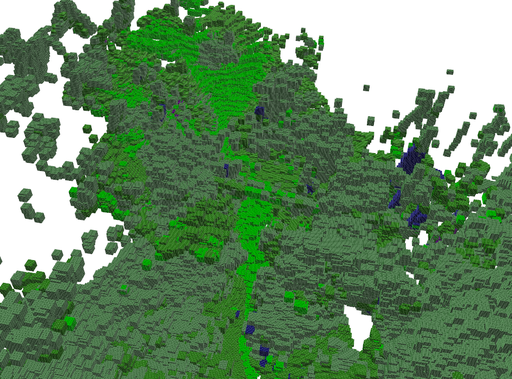}
        \label{fig:neya_trees_semantic2}
    \end{subfigure}  
    \hfill
    \begin{subfigure}[t]{0.24\textwidth}
        \centering
        \includegraphics[width=\textwidth]{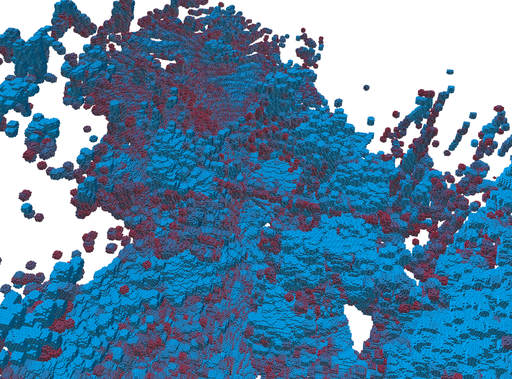}
        \label{fig:neya_trees_variance2}
    \end{subfigure}
    \hfill
        \begin{subfigure}[t]{0.24\textwidth}
        \centering
        \includegraphics[width=\textwidth]{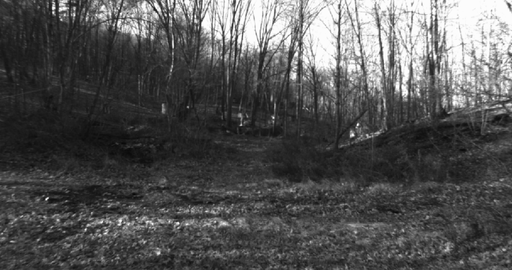}
        \label{fig:neya_cam_building2}
    \end{subfigure}
    \hfill
    \begin{subfigure}[t]{0.24\textwidth}
        \centering
        \includegraphics[width=\textwidth]{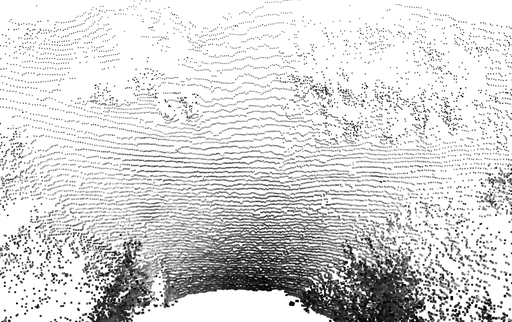}
        \label{fig:neya_pc_building2}
    \end{subfigure}  
    \hfill
    \begin{subfigure}[t]{0.24\textwidth}
        \centering
        \includegraphics[width=\textwidth]{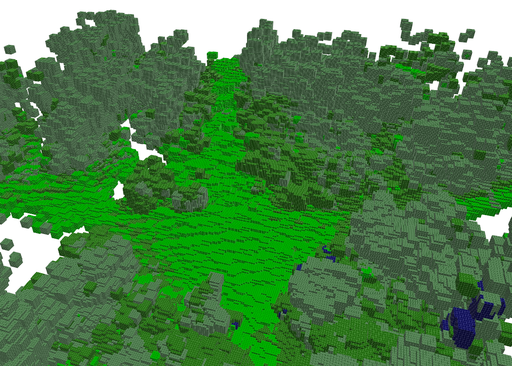}
        \label{fig:neya_building_semantic2}
    \end{subfigure}  
    \hfill
    \begin{subfigure}[t]{0.24\textwidth}
        \centering
        \includegraphics[width=\textwidth]{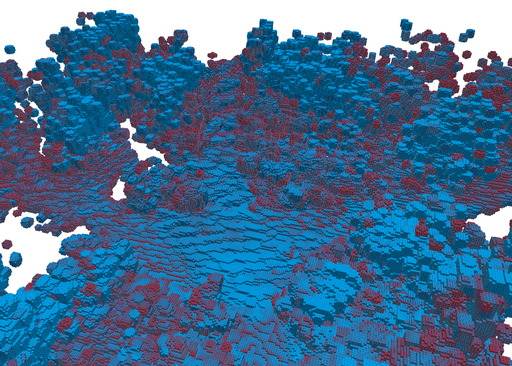}
        \label{fig:neya_building_variance2}
    \end{subfigure}
    \hfill
    \begin{subfigure}[t]{0.24\textwidth}
        \centering
        \includegraphics[width=\textwidth]{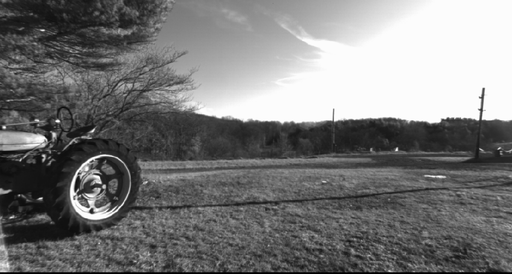}
        \label{fig:neya_cam_chall}
    \end{subfigure}
    \hfill
    \begin{subfigure}[t]{0.24\textwidth}
        \centering
        \includegraphics[width=\textwidth]{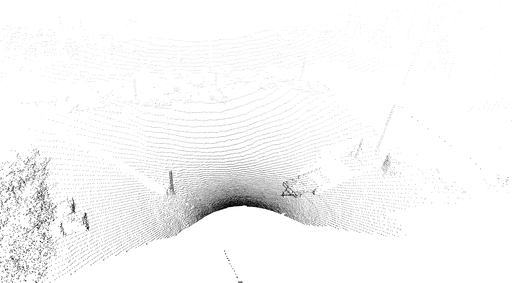}
        \label{fig:neya_pc_chall}
    \end{subfigure}  
    \hfill
    \begin{subfigure}[t]{0.24\textwidth}
        \centering
        \includegraphics[width=\textwidth]{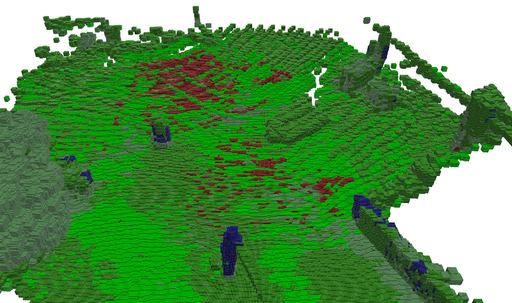}
        \label{fig:neya_chall_semantic}
    \end{subfigure}  
    \begin{subfigure}[t]{0.24\textwidth}
        \centering
        \includegraphics[width=\textwidth]{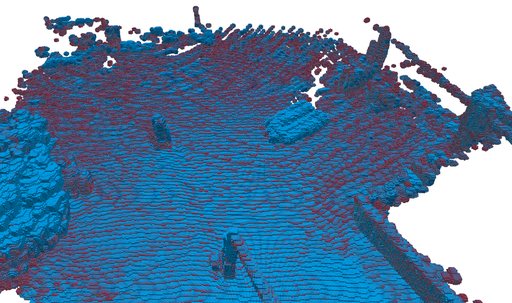}
        \label{fig:neya_chall_variance}
    \end{subfigure}
    \caption{Raw data and semantic maps from off-road testing. Each row contains a different scene from testing. From top to bottom, the scenes show a building with telephone poles, dense forest, a narrow path through forest, another narrow path through dense forest, and lastly a challenging scene with a tractor, building, and telephone poles. Each row displays the raw sensor data from the camera and point cloud for reference, as well as the semantic and variance maps. }
    \label{fig:neya_data2}
\end{figure*}

Data was collected with support from Neya Systems at two of their off-road test facilities with a variety of localization and perception challenges. The first test facility was a large outdoor field with few semantic categories other than bushes and trees. The second, more challenging test facility, was an outdoor paintball facility containing pathways covered by fallen leaves, bare trees, steep hills, buildings, mud, telephone poles, and other difficult features. We recorded several ROS bags over the span of three days of testing. Semantic maps and variance maps were constructed in ROS with the poses from LIO-SAM and filtered front LiDAR. 

Images of the raw data as well as semantic and variance maps are shown in Fig. \ref{fig:neya_data1} and \ref{fig:neya_data2}. Fig. \ref{fig:neya_data1} includes higher resolution images to demonstrate the sparsity of the data and more vividly portray the maps. In the scene, the vehicle drives along a path with two large buildings on the left, trees and bushes on the right, and another patch of foliage in front. Despite never encountering the building class, the buildings are labeled with reasonable guesses of mostly tree (brown), with patches of grass (green) or bush (dark green) on the horizontal portions and pedestrian (blue) on the narrow vertical portions. Buildings most closely resemble trees, which are also large un-traversable obstacles. The regions of the large building classified as pole, grass, or bush have high variance and are likely labeled as such due to the sparsity of the point cloud demonstrated in the top right of Fig. \ref{fig:neya_data1}. At this point, the robot has observed few points from the building and is more prone to making misclassification errors. Other areas with high variance include dense foliage, again due to the difficulty of labeling sparse point cloud data. In particular, voxels at the boundaries between different categories, such as bush and tree or bush and grass, have high variance. The driveable surface is generally segmented correctly with low variance as grass with some bushes and rubble. 

Several more examples from our dataset are shown in Fig. \ref{fig:neya_data2}. In order to include more frames, the images are at a lower resolution. Each row demonstrates a different example scene from our data, with the raw camera image, point cloud, semantic map, and variance map from left to right. The first row shows an open plain with telephone poles, trees to the sides, and a large building in front. The telephone pole is correctly classified as a pole with low variance, and the building in front is again classified as a tree, with the roof classified as bush. The driveable surface is segmented as grass with some patches of rubble (red) where there are rocks in the path. The patches of rubble have high variance due to sparse point clouds and few rocks, as the network mixes predictions between rubble and grass. Other regions of high variance include the boundaries of dense foliage between bushes and trees.

The next row demonstrates a dense forest scene with tall trees, bushes, and no easily segmented path. The tall trees are sparsely represented in the point cloud, creating mis-classifications of trees as poles or bushes. Sparsely covered trees have high uncertainty in the variance map. In contrast, the driveable surface of grass and some bushes is easily segmented and has low uncertainty. 

The next two rows of images show the vehicle traversing a narrow path through hills surrounded by trees and bushes. Despite the difficult terrain, the driveable surface is correctly segmented with low uncertainty. The dense foliage has high variance at the boundaries of the path and bushes, as well as at mis-classifications of trees as poles. 

The final row contains a difficult scene with telephone poles and wires, a tractor, buildings, and more challenging terrain. To the left of the map is a forested region with a tractor to the right side, which is labeled in the semantic map as a cluster of pole (blue), bush, grass and tree, since the semantic segmentation network is unable to identify the tractor. The raw point cloud of the tractor displays how the wheels may be interpreted as poles or people, and the body of the tractor as bush. In the variance map, the tractor has high uncertainty. The ground is more difficult for the semantic segmentation network to label due to rocks and gravel, resulting in mixed rubble and grass labels in the semantic map and high variance in the variance map. Also included in the map to the right is a small trailer from a truck which is labele as a bush, and telephone poles with telephone wires. The poles are labeled as pedestrians or trees, while the wires are classified as bushes. The telephone poles and wires have high uncertainty since they are not captured in the training set. 

% A video of ConvBKI mapping on ROS bags from real-world testing is available in the supplementary material, and a sample ROS bag is available on GitHub. Overall, ConvBKI is capable of successfully generating maps with meaningful labels despite perceptual challenges, including sparse point clouds, semantic categories outside the training set, and transferring between data sets. By operating explicitly within a differentiable probabilistic framework, ConvBKI maintains reliability on challenging data with quantifiable uncertainty. Additionally, ConvBKI leverages the existing decomposable robotics pipeline, where features are produced from sensor data and recurrently integrated into a map. In contrast, purely implicit deep learning approaches which hallucinate entire scenes can fail unpredictably when exposed to data different from the training set due to a lack of structure and a large number of trainable parameters. 
\section{Conclusion}\label{sec:Conclusion}
In this paper, we introduced a differentiable 3D semantic mapping algorithm that combines the reliability and trustworthiness of classical probabilistic mapping algorithms with the
efficiency and differentiability of modern neural networks. We quantitatively compared efficiency and accuracy with probabilistic approaches, as well as reliability against a purely implicit deep learning method. To further study the resilience of our model on perceptually degraded driving scenarios, we gathered a new off-road dataset with semantic categories the network had not been trained on and studied the semantic and variance maps. 

{\color{black}Overall, ConvBKI is capable of successfully generating maps with meaningful labels despite perceptual challenges, including sparse point clouds, semantic categories outside the training set, and transferring between data sets. By operating explicitly within a differentiable probabilistic framework, ConvBKI maintains reliability on challenging data with quantifiable uncertainty. Additionally, ConvBKI leverages the existing decomposable robotics pipeline, where features are produced from sensor data and recurrently integrated into a map. In contrast, purely implicit deep learning approaches which hallucinate entire scenes can fail unpredictably when exposed to data different from the training set due to a lack of structure and a large number of trainable parameters.} 

\textbf{Limitations of ConvBKI:} While ConvBKI achieves quick inference rates and expressive, automatically tuned kernels, it has several noteworthy limitations which encourage future work. First, ConvBKI is computationally efficient and parallelizable due to the depthwise convolution operation which applies the same discretized kernel to each semantic category. While the convolution operation accelerates performance, it does not model rotation of points which may be useful in sloped or less structured environments. Second, kernel optimization performs best when the provided training data is similar to the testing data. Training on over-fitted segmentation predictions results in decreased spatial smoothing since the predictions are already near perfect, while training on noisy segmentation predictions results in more expressive spatial kernels. Third, as previously discussed, ConvBKI does not consider the motion of dynamic objects and leaves artifacts in the map.

% \textbf{Limitations of ConvBKI:}
% \begin{enumerate}
%     \item Rotation of objects
%     \item Training on over-fitted networks
% \end{enumerate}

\textbf{Future Work:} Several avenues for future work exist, and we hope our open source software will encourage other roboticists to extend our method. First, our method can be applied to mesh-based semantic mapping algorithms such as Hydra \cite{Hydra} by replacing the semantic update with ConvBKI. ConvBKI can accelerate the framework and provide voxel-wise variance calculations which may be used to estimate uncertainty at the object level. Another extension is to remove the traces of dynamic objects through a well-formulated network layer which operates explicitly on the Dirichlet distribution concentration parameters such as in Dynamic BKI \cite{DynamicBKI}. {\color{black}Third, while we tested ConvBKI on both camera and LiDAR in this paper, optimal fusion of the two sensors in the BKI framework remains an open question. Two possibilities are to separately learn kernels for each sensor and add the posterior maps, or to fuse the dense semantic detail from cameras with the accurate geometry of LiDAR before map inference. Finally, ConvBKI may benefit from more expressive extended likelihood distributions which consider the rotation of points instead of only the scale of each category.}

% the results of ConvBKI demonstrated that defining more expressive extended likelihood distributions over the input points through compound kernels improved the quantiative performance, suggesting that further increasing expressivity of the input may also improve the performance. ConvBKI is computationally efficient due to the depthwise convolution operation, but does not consider the rotation of input points. A future direction for ConvBKI is the incorporation of per-point extended likelihoods through learned per-point rotations and scaling factors.

%%%%%%%%%%%%%%%%%%%%%%%%%%%%%%
% \input{related_work}
% \input{drafts}
%%%%%%%%%%%%%%%%%%%%%%%%%%%%%%
% \clearpage
{%\small
% \balance
% \bibliographystyle{template/IEEEtran}
\bibliographystyle{IEEEtran}
\bibliography{bib/strings-abrv,bib/ieee-abrv,bib/refs}
}

\begin{IEEEbiography}[{\includegraphics[width=1in,height=1.25in,clip,keepaspectratio]{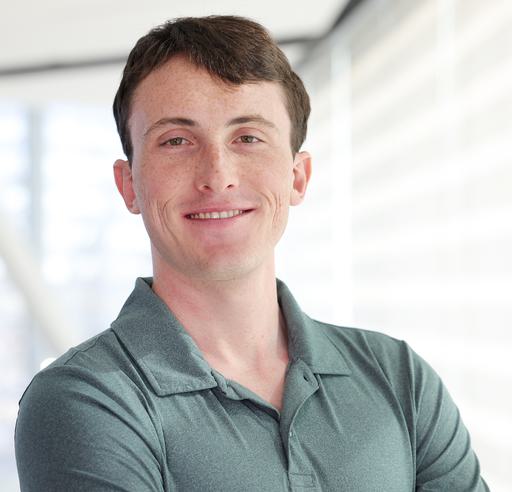}}]{Joey Wilson} received the B.S. degree in computer engineering from California Polytechnic State University San Luis Obispo (Cal Poly), CA, USA, in 2019. He is currently a Ph.D. candidate in the University of Michigan Robotics Department, Ann Arbor, MI, USA. His research interests include spatial memory in dynamic and uncertain environments for autonomous systems.
\end{IEEEbiography}

\begin{IEEEbiography}[{\includegraphics[width=1in,height=1.25in,clip,keepaspectratio]{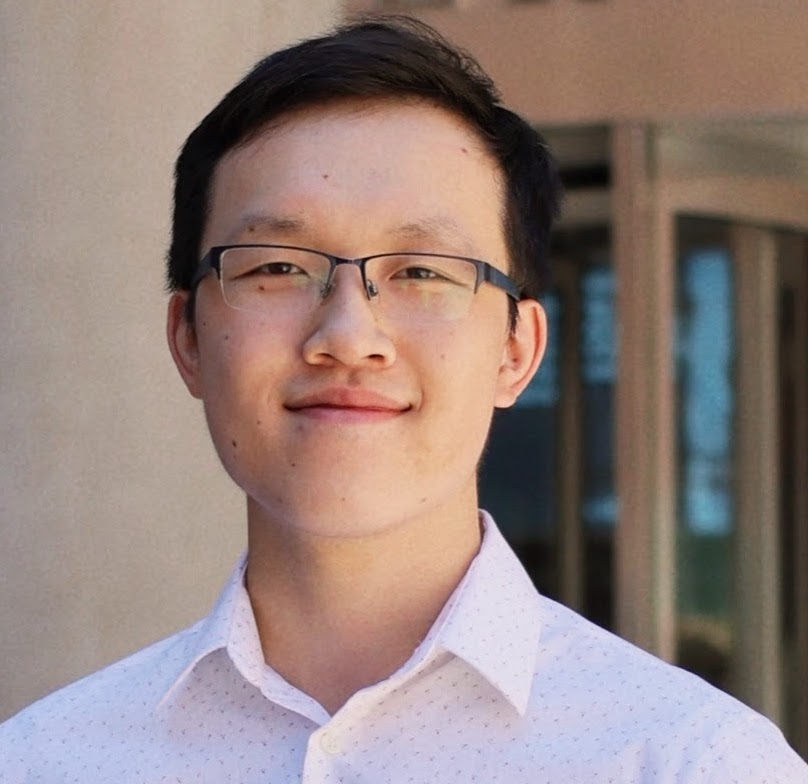}}]{Yuewei Fu} received the B.S. degree in mechanical engineering from New York University (NYU) in 2021. He is currently a M.S. student in the University of Michigan Robotics Department, Ann Arbor, MI, USA. His research interests include scene understanding for off-road and underwater robots in dynamic environments.
\end{IEEEbiography}

\begin{IEEEbiography}[{\includegraphics[width=1in,height=1.25in,clip,keepaspectratio]{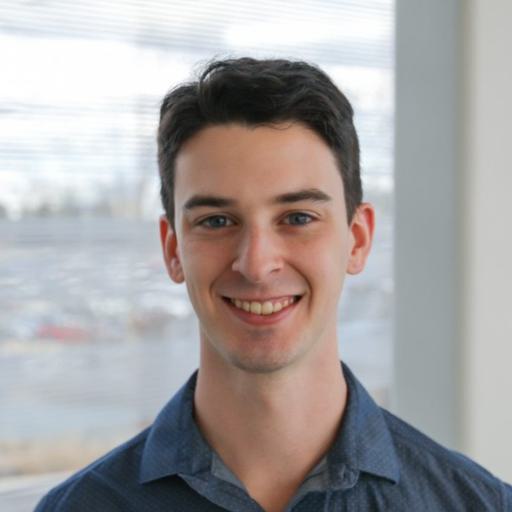}}]{Josh Friesen} received the B.S. degree in software engineering from Fresno Pacific University in 2021. He is currently a M.S. student in the University of Michigan Robotics Department, Ann Arbor, MI, USA. His research interests include deep learning for autonomous robots.
\end{IEEEbiography}

\begin{IEEEbiography}[{\includegraphics[width=1in,height=1.25in,clip,keepaspectratio]{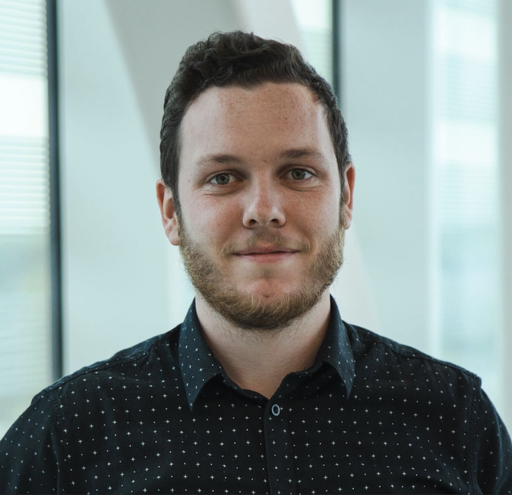}}]{Parker Ewen} received his MSc in Robotics, Systems, and Control from the ETH Zurich in 2020. He is currently a PhD candidate at the University of Michigan where he is a member of ROAHM Lab. He researches numerical techniques for state estimation, mapping, and active learning with applications for robotics.
\end{IEEEbiography}

\begin{IEEEbiography}[{\includegraphics[width=1in,height=1.25in,clip,keepaspectratio]{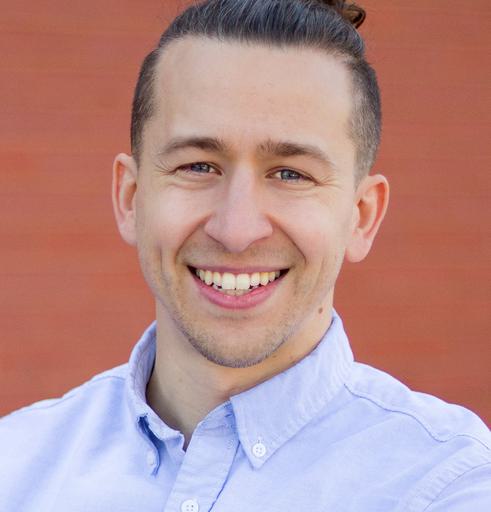}}]{Andrew Capodieci} is the Director of Robotics at Neya Systems and has been with Neya for over a decade. During this time, Andrew has performed and managed applied research in all areas of the robotics stack including kinodynamically-feasible path planning in congested spaces, traversability estimation in off-road terrain, and negative obstacle detection. As Director of Robotics, Andrew is focused on transitioning Neya's state-of-the-art autonomy research into fieldable, robust autonomy capabilities that deliver value to the warfighter and commercial off-road spaces. Andrew has led numerous multi-million-dollar programs including Neya's work on GVSC's Combat Vehicle Robotics program, and programs to develop autonomous construction vehicles for the commercial sector.
\end{IEEEbiography}

\begin{IEEEbiography}[{\includegraphics[width=1in,height=1.25in,clip,keepaspectratio]{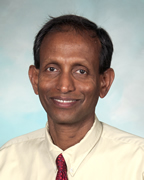}}]{Paramsothy~Jayakumar} received the B.Sc.Eng. degree (Hons.) from the University of Peradeniya, Sri Lanka, and the M.S. and Ph.D. degrees from Caltech. He worked at Ford Motor Company and BAE Systems. He is a Senior Technical Expert of analytics with U.S. Army DEVCOM Ground Vehicle Systems Center (GVSC). He has published over 200 papers in peer-reviewed literatures. He is a fellow of the Society of Automotive Engineers and the American Society of Mechanical Engineers. He received the DoD Laboratory Scientist of the Quarter Award, the NATO Applied Vehicle Technology Panel Excellence Awards, the SAE Arch T. Colwell Cooperative Engineering Medal, the SAE James M. Crawford Technical Standards Board Outstanding Achievement Award, the BAE Systems Chairman’s Award, and the NDIA GVSETS Best Paper Awards. He is also an Associate Editor of the ASME Journal of Autonomous Vehicles and Systems, and the Editorial Board Member of the International Journal of Vehicle Performance and the Journal of Terramechanics.
\end{IEEEbiography}

\begin{IEEEbiography}[{\includegraphics[width=1in,height=1.25in,clip,keepaspectratio]{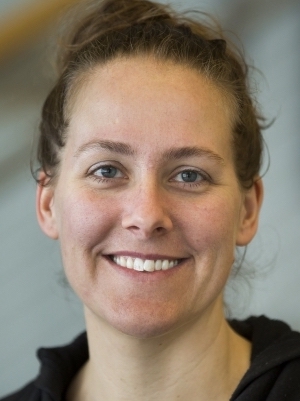}}]{Kira Barton} received the Ph.D. degree in Mechanical Engineering from the University of Illinois at Urbana-Champaign, USA, in 2010. She is currently a Professor at the Robotics Institute and Department of Mechanical Engineering, University of Michigan, Ann Arbor, MI, USA. Her research interests lie in control theory and applications including high precision motion control, iterative learning control, and control for autonomous vehicles.
\end{IEEEbiography}

\begin{IEEEbiography}
[{\includegraphics[width=1in,height=1.25in,clip,trim={3.1cm 0 3.1cm 3.5cm},keepaspectratio]{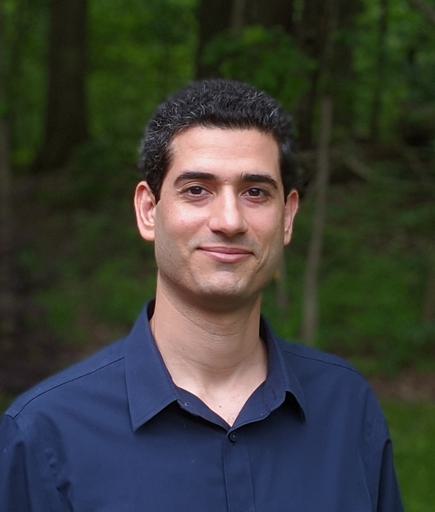}}]{Maani Ghaffari}
received the Ph.D. degree from the Centre for Autonomous Systems (CAS), University of Technology Sydney, NSW, Australia, in 2017. He is currently an Assistant Professor at the Department of Naval Architecture and Marine Engineering and the Department of Robotics, University of Michigan, Ann Arbor, MI, USA. He is the director of the Computational Autonomy and Robotics Laboratory. He is the recipient of the 2021 Amazon Research Awards. His research interests lie in the theory and applications of robotics and autonomous systems.
\end{IEEEbiography}

\end{document}